\documentclass{PERS}

\usepackage{lmodern}

\usepackage{natbib}
\usepackage{amsmath,amsfonts,amssymb,bm}

\usepackage{newtxtext}
\usepackage{newtxmath}

\usepackage{hyperref}  
\usepackage{caption}   
\usepackage{booktabs,tabularx}
\renewcommand{\figurename}{Figure}
\renewcommand{\tablename}{Table}
\usepackage{adjustbox}
\usepackage{booktabs}   
\usepackage{multirow}   
\usepackage{array}      
\usepackage{makecell}  
\usepackage{arydshln}      
\usepackage{setspace}  
\usepackage[symbol]{footmisc}

\nolinenumbers

\title{Multimodal Remote Sensing Image Registration: A Comprehensive Review, Challenges and Prospects}

\authors{Zhiqiang Han,  Yuanxin Ye, Qiuyun Wu, Jinhao Chen, Bai Zhu, Siyuan Hao \correspondingauthor{Siyuan Hao (syhao@bjtu.edu.cn)}}
\affiliation{Zhiqiang Han,  Yuanxin Ye, Qiuyun Wu and Jinhao Chen are with the Faculty of Geosciences and Engineering, Southwest Jiaotong University. Chengdu, 611756, China. \\
	Bai Zhu is with the Yunnan Key Laboratory of Quantitative Remote Sensing \slash\ Yunnan International Joint Laboratory for Integrated Sky-Ground Intelligent Monitoring of Mountain Hazards, Faculty of Land Resources Engineering, Kunming University of Science and Technology. Kunming, 650093, China. \\
	Siyuan Hao is with the School of Software Engineering, Beijing Jiaotong University. Beijing, 100044, China. \\
	Corresponding author: Siyuan Hao (syhao@bjtu.edu.cn)}
	
\abstract{Multimodal remote sensing image registration is a crucial prerequisite for the collaborative processing and downstream application of remote sensing data, such as image fusion, change detection, and target recognition. However, significant variations in radiometry, geometry, scale, viewpoint, and time often exist between multimodal images. These differences, driven by varying sensor geometries, physical radiation mechanisms, imaging platforms, and environmental disturbances, pose severe challenges to achieving high-precision, robust registration.\\This paper systematically reviews the progress of mainstream multimodal remote sensing image registration methods. Based on their registration pipelines, existing approaches are categorized into three main types: region-based, feature-based, and deep learning-based methods. We detail the core principles, representative algorithms, advantages, and limitations of each category. Additionally, we summarize publicly available multimodal image datasets in the remote sensing domain, analyzing their specific characteristics and applicable scenarios. Finally, we highlight current bottlenecks in high-precision registration research and outline future development trends. This review aims to provide a comprehensive reference and valuable insights for researchers in related fields.}

\begin{document}
	\maketitle
	
\noindent
\textbf{Table of Contents Description:} This paper presents a comprehensive review of multimodal remote sensing image registration, systematically categorizing region-based, feature-based, and deep learning-based paradigms while analyzing mainstream datasets, cutting-edge challenges, and future development trends.
\vspace{2em}

\footnotetext[0]{This paper is supported by the National Key Research and Development Program of China under Grant 2024YFC3015404, in part by the National Natural Science Foundation of China under Grant 42271446, and in part by the Science and Technology Program of Tianjin under Grant 24YFYSHZ00080.}

%
\section{Introduction}
\begin{figure}[ht]
	\centerline{\includegraphics[width=\linewidth]{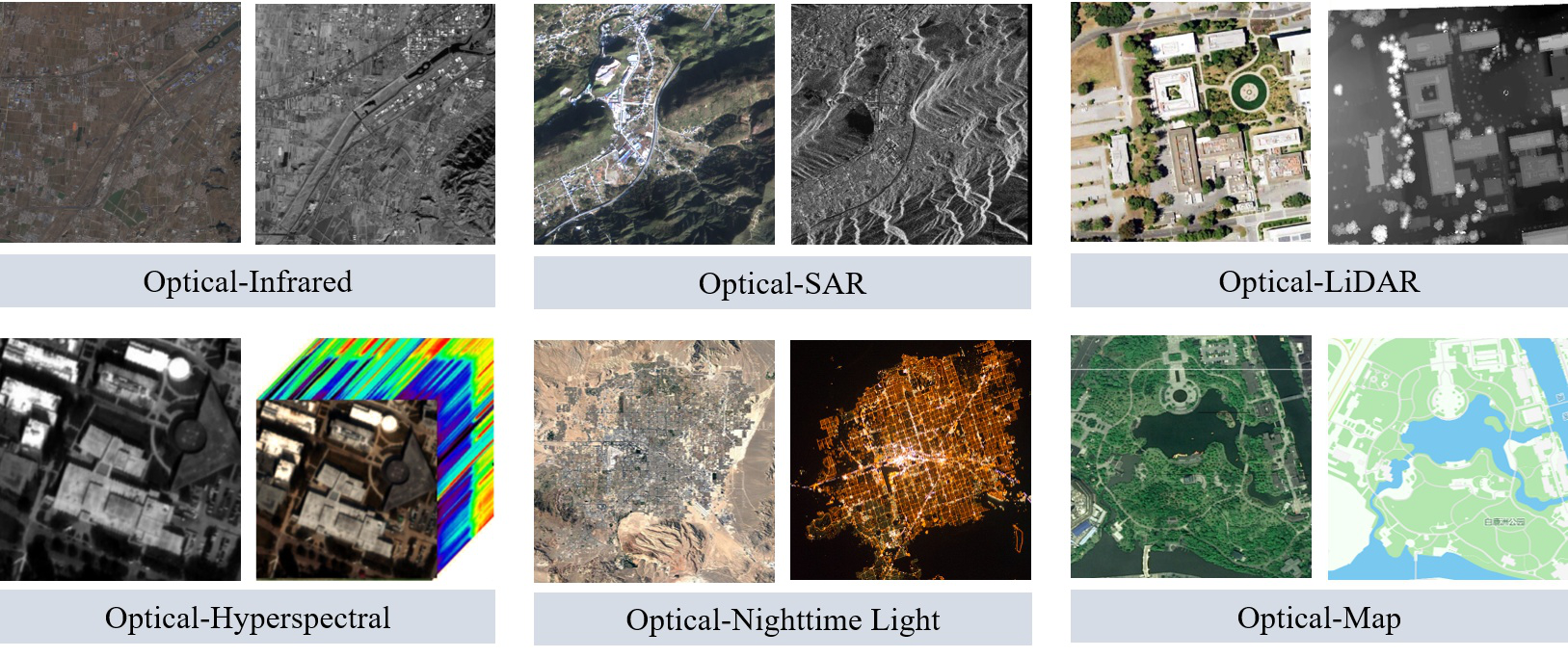}}
	\caption{Multimodal remote sensing image pairs.}
	\label{Fig:1}
\end{figure}

Multimodal remote sensing images are typically obtained from observations of the same object at different times using different sensors and imaging mechanisms, such as optical-infrared image \citep{li_review_2022}, optical-hyperspectral image \citep{wang_hyperspectral_2024} , optical-SAR image \citep{kulkarni_pixel_2020}, optical-LiDAR image \citep{wang_imbalance_2023}, optical-nighttime light image and optical-map image \citep{jiang_review_2021}. As shown in \autoref{Fig:1}.

Compared with single-modal sensors, the joint processing of multimodal remote sensing data yields more reliable, comprehensive, and accurate observation results \citep{brown_survey_1992}. Consequently, research on multimodal image registration is of significant scientific importance. Furthermore, registration serves as a fundamental prerequisite for numerous downstream applications, such as image fusion \citep{tang_deep_2023}, change detection \citep{sun_similarity_2024}, and object detection \citep{belmouhcine2023multimodal}. Ultimately, effective multi-level and multifaceted Earth observation relies on the thorough integration of these diverse data sources \citep{tang_divfusion_2023}.

\begin{figure}[ht]
	\centerline{\includegraphics[width=\linewidth]{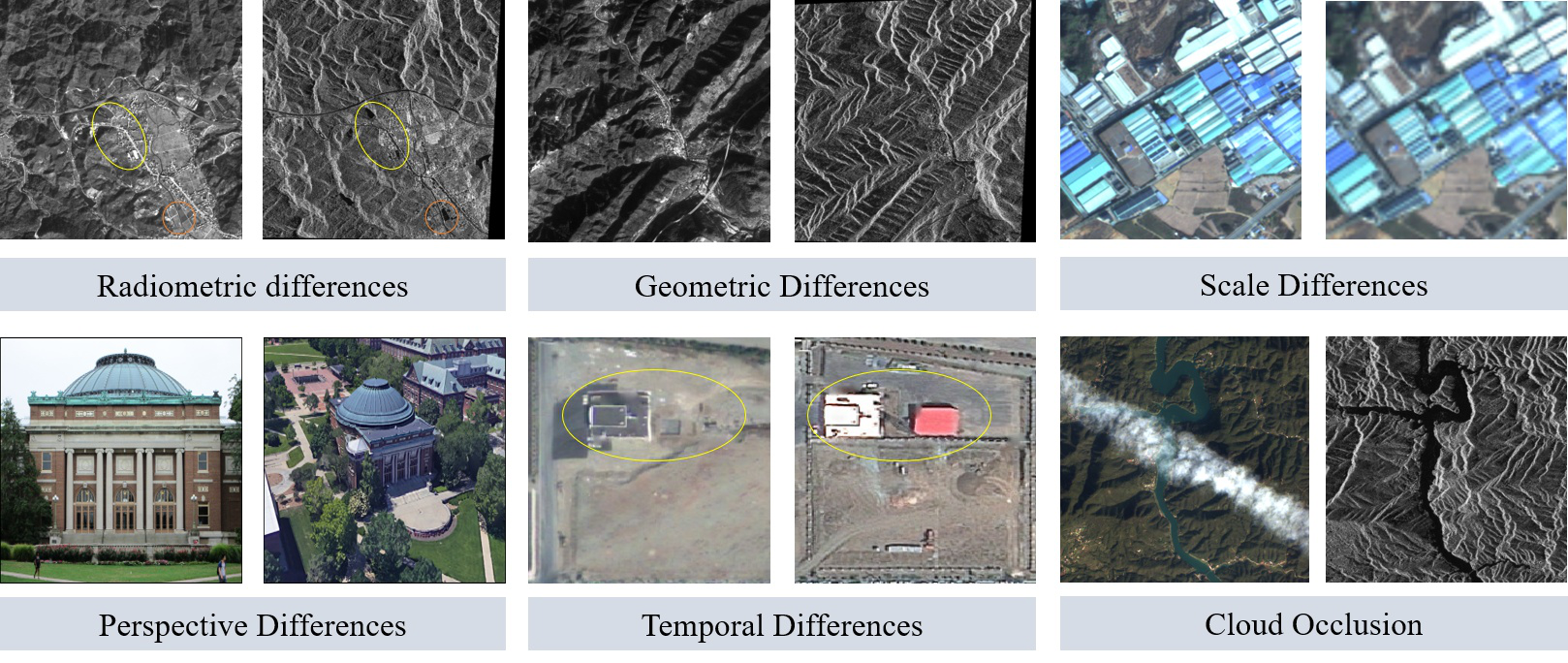}}
	\caption{Challenges in multimodal remote sensing image registration.}
	\label{Fig:2}
\end{figure}

Multimodal remote sensing images inherently exhibit significant radiometric, geometric, scale, perspective, and temporal discrepancies. These arise from the distinct physical radiation mechanisms and spatial configurations of various sensors (e.g., infrared, LiDAR, and SAR) \citep{ye_robust_2022,zhu_robust_2021}, as well as the diverse platforms (satellite, aerial, and ground) that host them. Furthermore, image quality is highly susceptible to environmental and meteorological conditions \citep{feng_registration_2021}. Consequently, pervasive interferences such as noise, cloud occlusion, and blur pose substantial challenges to high-precision image registration. \autoref{Fig:2} illustrates these primary challenges.

The primary objective of multimodal remote sensing image registration is to accurately identify uniformly distributed corresponding point pairs across images to derive an optimal geometric transformation model \citep{xiang_global--local_2023}. Let $I_1(x, y)$ and $I_2(x, y)$ denote the reference and input images, respectively. Their geometric mapping relationship can be formulated as $I_2(x, y) = g\left[I_1\left(f(x, y)\right)\right]$, where $f$ represents the 2D geometric transformation function and $g$ denotes the 1D grayscale transformation function \citep{brown_survey_1992}.

Traditional registration methods are primarily suited for single-modal images with linear radiometric differences. When applied to multimodal remote sensing images exhibiting significant geometric distortions and nonlinear radiometric variations, these methods struggle to construct robust feature descriptors and similarity measures capable of capturing shared characteristics. Consequently, although mainstream commercial software (e.g., ENVI, ERDAS, and PCI) provides basic registration tools, their reliance on traditional techniques prevents high-precision multimodal registration. To address these limitations, recent research has focused on developing highly accurate, efficient, and robust multimodal registration methods.

\begin{figure}[htb]
	\centerline{\includegraphics[width=\linewidth]{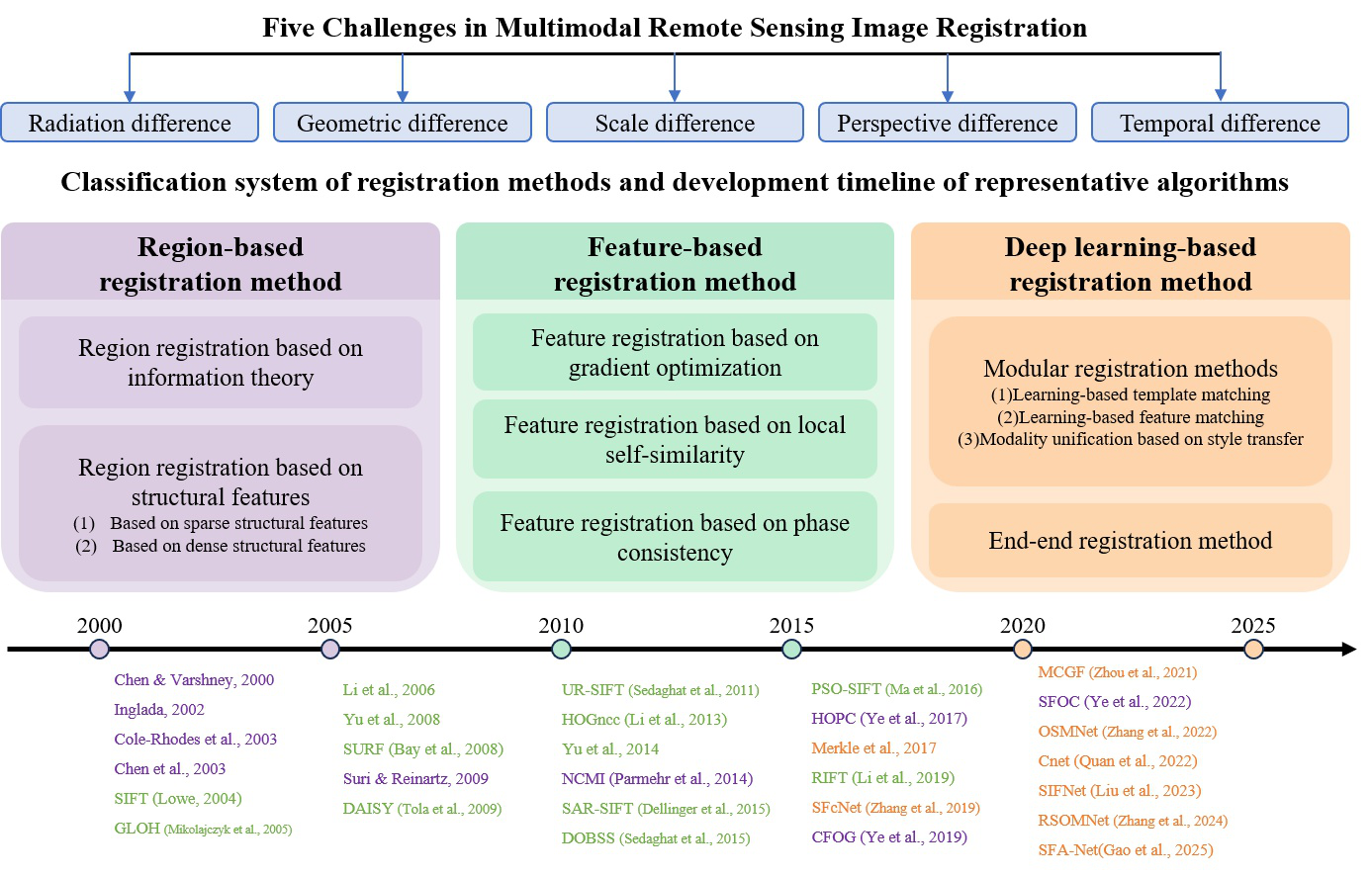}}
	\caption{Overall framework of this survey.}
	\label{Fig:3}
\end{figure}

Based on distinct registration processes, traditional methods are primarily classified into region-based and feature-based approaches \citep{sommervold_survey_2023}. Recently, thanks to their robust ability to extract high-dimensional representations, deep learning algorithms have firmly established themselves as a crucial third pillar in multimodal remote sensing registration \citep{jiang_systematic_2015}. Guided by this taxonomy, this paper systematically reviews the current research landscape. The overall structure is illustrated in \autoref{Fig:3}. First, we provide a detailed review tracing the evolution of these three methodological categories from single-modal to multimodal applications, discussing representative algorithms and providing links to open-source implementations. Subsequently, we present a comprehensive analysis of publicly available multimodal datasets. Finally, we conclude by highlighting current bottlenecks in high-precision registration and outlining future research directions.

\section{Region-based registration method}
Region-based registration methods, often referred to as template-based methods, fundamentally rely on optimizing an appropriate similarity metric to accurately estimate geometric transformation parameters. Unlike natural images, a distinct advantage of remote sensing imagery is the availability of precise prior geometric information, such as Position and Orientation System (POS) or Rational Polynomial Coefficient (RPC) data \citep{shen_correcting_2017,wu_automatic_2016}. Leveraging this georeferencing data effectively mitigates significant scale and rotation discrepancies between multimodal images. Because the initial georeferencing error is typically limited to a few dozen pixels in the image space, it provides a reliable coarse search range for subsequent template matching \citep{jiang_systematic_2015}.

Based on the domain of image representation, region-based registration algorithms can be broadly divided into spatial and frequency categories. Within spatial domain techniques, representative similarity metrics typically include the sum of squared differences (SSD), normalized cross-correlation (NCC), and mutual information (MI) \citep{amin_remote_2015,chen_mutual_2003}. In contrast, approaches operating in the frequency domain primarily rely on phase correlation (PC) as their core measure \citep{tong_novel_2015}. Because PC exploits the inherent scaling, rotational, and translational properties of the Fourier transform, it demonstrates strong capability in handling these geometric discrepancies across images \citep{reddy_fft-based_1996}.

However, the aforementioned traditional region-based methods typically drive the registration process by optimizing intensity-based similarity metrics to estimate geometric transformations. Consequently, approaches like SSD, NCC, and PC fail to yield robust results for multimodal remote sensing images, as these metrics are highly sensitive to the severe nonlinear radiometric variations inherent in such data \citep{yang_adjacent_2024}.

Inspired by medical imaging applications \citep{wells_multimodal_1996}, information-theoretic metrics, particularly mutual information (MI), were the first widely adopted similarity measures for multimodal remote sensing registration. To further improve accuracy and computational efficiency, subsequent research introduced a series of novel techniques, defined herein as structure-feature-based methods. The following sections review these two categories in detail.

\subsection{Region registration method based on information theory}
Leveraging the statistical correlation between images, information-theoretic methods have been widely applied to multimodal remote sensing registration \citep{chen_pyramid_2000}. By effectively accommodating non-linear radiometric variations, these approaches demonstrate significantly greater robustness than traditional metrics such as SSD, NCC, and PC. Building upon this foundation, numerous enhanced methods—most notably mutual information (MI) \citep{chen_mutual_2003} and its advanced variants \citep{cole-rhodes_multiresolution_2003,suri_mutual-information-based_2010, xu_multimodal_2016}—have been developed for diverse multimodal tasks. Finally, \autoref{Table:1} summarizes the core concepts of representative region-based registration algorithms.

\begin{table}[htbp]
	\centering
	\renewcommand{\arraystretch}{1.3}  
	\caption{Region registration method based on information theory}
	\label{Table:1}
	\begin{tabular*}{\textwidth}{@{\extracolsep{\fill}} 
			>{\centering\arraybackslash}m{5cm} 
			>{\centering\arraybackslash}m{11cm} @{}}
		\toprule
		Representative algorithm & Core ideas \\
		\midrule
		\cite{chen_pyramid_2000} & Image pyramids \& MI; Robust to nonlinear radiative differences \\
		\cite{inglada_similarity_2002} & Evaluation of MI; Good comprehensive performance \\
		\cite{chen_mutual_2003} & GPVE optimization; Solves interpolation artifacts \\
		\cite{cole-rhodes_multiresolution_2003} & SPSA gradient policy optimization; Accelerated convergence \\
		\cite{suri_mutual-information-based_2010} & Feature selection for high-resolution images; Improved performance \\
		NCMI \citep{parmehr_automatic_2014} & Histogram-based non-parametric PDF estimation; Improved performance \\
		\cite{xu_multimodal_2016} & Symmetric Kullback-Leibler divergence; Robust to insufficient overlap \\ 
		TOMI \citep{li_hybrid_2021} & Tensor direction parallelism \& gradient MI; Robust to nonlinear radiation \\
		\cite{wangFastRegistrationHighResolution2024} & Grayscale NMI ; A novel segmentation strategy \\
		\bottomrule
	\end{tabular*}
\end{table}

\cite{chen_pyramid_2000} introduced a wavelet-based pyramid registration method, providing early evidence of MI's potential in multimodal remote sensing applications. Subsequently, \cite{inglada_similarity_2002} evaluated four similarity metrics using multi-temporal SPOT imagery, concluding that MI delivers superior overall performance despite radiometric dissimilarities. Furthermore, \cite{chen_mutual_2003} investigated interpolation algorithms for MI joint histogram estimation, identifying nearest-neighbor interpolation as the most effective. To mitigate interpolation-induced artifacts during MI calculations, they also proposed generalized partial volume estimation (GPVE).

To enhance MI optimization, a gradient-based simultaneous perturbation stochastic approximation (SPSA) approach was utilized by \cite{cole-rhodes_multiresolution_2003}, achieving successful alignment of multi-sensor and multi-temporal image pairs. Focusing on complex urban environments at high resolutions, an automated hierarchical (coarse-to-fine) MI pipeline was constructed by \cite{suri_mutual-information-based_2010} specifically for matching Ikonos and TerraSAR-X datasets. To tackle the alignment of airborne LiDAR with optical data, \cite{parmehr_automatic_2014} presented the normalized combined mutual information (NCMI) metric, which substantially boosted registration precision. Additionally, the Jeffrey divergence—functioning as a symmetric variation of the Kullback-Leibler divergence—was put forward by \cite{xu_multimodal_2016}, proving its advantages compared to previously modified MI techniques. Later, \cite{li_hybrid_2021} established the tensor orientation and mutual information (TOMI) measure. By fusing gradient distributions with structural orientations, the TOMI framework delivers superior resilience against heavy noise and extreme nonlinear radiometric shifts. Most recently, to overcome the ongoing bottlenecks of computational inefficiency and local distortions in high-resolution scenarios, \cite{wangFastRegistrationHighResolution2024} introduced a rapid automated technique driven by grayscale normalized mutual information (GNMI). Through an innovative block-based segmentation scheme that merges varying patch sizes, this strategy bypasses conventional feature extraction, thereby concurrently elevating processing speed and alignment accuracy for strongly heterogeneous SAR-optical image pairs.

Overall, information-theoretic methods have proven highly effective in handling multimodal registration across diverse imaging mechanisms by relying on joint histogram-based statistical similarities. However, they have traditionally suffered from significant limitations, primarily their high computational complexity, which makes them extremely time-consuming. Although recent advancements—such as the aforementioned block-based segmentation strategies—have begun to successfully mitigate these efficiency bottlenecks, the inherent computational demands have historically restricted their widespread application, particularly for large-scale multimodal imagery.

\subsection{Structure-feature-based region registration methods}
Depending on whether feature extraction is pixel-wise, this review categorizes structure-feature-based methods into sparse and dense approaches. To establish reliable correspondences across multimodal images, these methods employ classical metrics to evaluate the similarity of the extracted features. Specifically, sparse structure-based methods typically capture geometric structures using a relatively sparse sampling grid \citep{li_matching_2013,xiong_rank-based_2020,ye_robust_2017}. Finally, \autoref{Table:2} summarizes representative structure-feature-based algorithms, detailing their core concepts and providing open-source code links where available.

\begin{table}[htbp]
	\centering
	\renewcommand{\arraystretch}{1.3}  
	\caption{Structural feature-based registration methods}
	\label{Table:2}
	\begin{tabular*}{\textwidth}{@{\extracolsep{\fill}} 
			>{\centering\arraybackslash}m{1.2cm} 
			>{\centering\arraybackslash}m{4cm} 
			>{\centering\arraybackslash}m{8cm} 
			>{\centering\arraybackslash}m{2cm} @{}}
		\toprule
		Method type & Representative algorithm & Core ideas & Acquisition channels \\
		\midrule
		\multirow{8}{*}[2.5ex]{\rotatebox{270}{Sparse structural features}} 
		& HOG\textsubscript{NCC} \citep{li_matching_2013} & HOG \& NCC combination & - \\
		& DLSS \citep{ye_robust_2017} & Dense local self-similar description; Shape attributes & \href{https://www.dropbox.com/s/o6f7j0igdiadlvs/DLSC\%20publish\%20code.rar?dl=0}{[Code]} \\
		& HOPC \citep{ye_robust_2017-1} & Histogram of Oriented Phase Consistency; Global satellite image processing & \href{https://github.com/yeyuanxin110/HOPC}{[Code]} \\
		& HOMPC \citep{fu_hompc_2018} & Log-Gabor filtering; Local amplitude \& phase consistency & - \\
		& RLSS \citep{xiong_rank-based_2020} & Spearman rank correlation \& LSS; Local shape attribute description & - \\
		\midrule 
		\multirow{11}{*}[2ex]{\rotatebox{270}{Dense structural features}} 
		& CFOG \citep{ye_fast_2019} & Pixel-by-pixel structural matching; HOG, LSS \& CFOG framework & \href{https://github.com/yeyuanxin110/CFOG}{[Code]} \\
		& OS-PC \citep{xiang_os-pc_2020} & Multi-channel feature representation; Fast optical-SAR registration & - \\
		& AWOG \citep{fan_exploiting_2021} & Angle-weighted strategy; CFOG gradient distribution & - \\
		& DCF-EPW \cite{xiang_robust_2022} & Dilated convolution features; Directional phase correlation & - \\
		& SFOC \citep{ye_robust_2022} & Multi-scale 1st \& 2nd gradients; Coarse-to-fine registration system & \href{https://github.com/yeyuanxin110/SFOC-Multimodal_Remote_Sensing_Image_Registration_System}{[Code]} \\
		& MAGD \citep{xiang_global--local_2023} & Multi-directional anisotropic Gaussian derivatives; High-accuracy evaluation dataset & \href{https://github.com/xym2009/OS-Eval}{[Code]} \\										
		\bottomrule
	\end{tabular*}
\end{table}

\cite{li_matching_2013} utilized the Histogram of Oriented Gradients (HOG) to represent structural features, employing zero-mean NCC as a similarity measure for SAR and optical image registration. Addressing shape attributes, \cite{ye_robust_2017} proposed the Dense Local Self-Similarity (DLSS) descriptor, leveraging NCC to achieve high-precision optical-to-SAR matching. Inspired by HOG, \cite{ye_robust_2017-1} extended the phase congruency (PC) model to capture structural attributes, introducing the Histogram of Oriented Phase Congruency (HOPC). When paired with NCC, HOPC demonstrated robust performance across optical, LiDAR, and SAR imagery exhibiting severe nonlinear radiometric differences, and has since been successfully scaled for global satellite image processing \citep{zhang_generalized_2021}. Building upon this foundation, \cite{fu_hompc_2018} developed a descriptor that integrates multi-directional amplitude and  from log-Gabor filters, further enhancing registration robustness. Additionally, \cite{xiong_rank-based_2020} designed the rank-based local self-similarity (RLSS) descriptor using the Spearman rank correlation coefficient, effectively characterizing local shape properties to achieve superior matching accuracy.

However, the aforementioned descriptors (e.g., HOG, HOPC, and DLSS) rely on relatively sparse sampling grids to capture structural features. This inherent sparsity inevitably leads to the loss of critical structural information, thereby limiting their registration accuracy and robustness. Conversely, dense structure-based methods employ pixel-wise descriptors to map image intensities into high-dimensional feature spaces, effectively driving the registration process \citep{fan_exploiting_2021, xiang_algorithm_2022, ye_fast_2019}.

\cite{ye_fast_2019} proposed a fast and robust template matching framework utilizing novel pixel-wise channel features of oriented gradients (CFOG). To maximize computational efficiency, they accelerated the sum of squared differences (SSD) optimization in the frequency domain via the fast Fourier transform (FFT). Notably, this highly adaptable framework can seamlessly integrate other pixel-level descriptors, such as HOG, LSS, and SURF. Addressing the failure of traditional phase correlation under severe nonlinear radiometric and geometric discrepancies, \cite{xiang_os-pc_2020} formulated specialized multi-channel feature representations for optical and SAR imagery. By employing the Sobel and ratio of exponentially weighted averages (ROEWA) operators, they developed two robust strategies to enhance 3D phase correlation performance across both spatial and frequency domains.

\cite{fan_exploiting_2021} enhanced CFOG via an angle-weighted strategy, distributing gradient values along the two principal directions to improve matching performance. To address large georeferencing errors, \cite{xiang_robust_2022} proposed the DCF-EPW (Dilated Convolutional Feature-Epipolar Oriented Matching Window) method. Its depthwise separable dilated convolutions (DCF) leverage large receptive fields to maintain positional invariance, while its epipolar-oriented rectangular templates capture more overlapping regions than traditional square templates, thereby establishing higher-precision optical-SAR correspondences.

Identifying that previous dense methods overlook local pixel relationships, multi-scale contexts, and superior second-order gradient details, \cite{ye_robust_2022} developed the SFOC (Steerable Filters of First- and Second-Order Channels) descriptor. By integrating both gradient orders, SFOC extracts highly discriminative structural features. Furthermore, they constructed an open-source, coarse-to-fine registration framework by accelerating traditional NCC via FFT and integral images. Finally, \cite{xiang_global--local_2023} formulated the multi-directional anisotropic Gaussian derivatives (MAGD) feature. Their approach successfully handles geometric distortions by pairing MAGD with a geometry-invariant mask within a confidence-guided local matching process. They also released a comprehensive evaluation dataset comprising hundreds of multimodal image pairs. Extending these multi-channel gradient principles to extreme environments, recent research has also integrated multi-channel oriented gradient templates with self-supervised super-resolution to tackle highly complex underwater terrain navigation in Arctic waters \citep{huangArcticUnderwaterTerrain2026}.

Within structure-feature-based approaches, sparse methods are restricted to calculations on predefined sampling grids, limiting their similarity evaluation entirely to the spatial domain. Consequently, frequency-domain metrics are inapplicable. In contrast, dense structure-based methods can seamlessly exploit similarity metrics in both the spatial and frequency domains. Balancing robustness and computational efficiency, these dense approaches demonstrate clear advantages in overcoming severe nonlinear radiometric variations, making them highly suitable for modern multimodal remote sensing applications.

 \section{Feature-based registration method}
 In summary, region-based methods are highly effective at overcoming radiometric differences when the initial spatial offset is limited. However, they remain highly sensitive to significant rotation and scale variations, and their robustness is severely compromised if prior georeferencing data is missing or highly inaccurate. To overcome these geometric constraints, feature-based registration methods provide a compelling alternative. Unlike region-based approaches that evaluate similarity over predefined templates or grids, feature-based methods generally do not require georeferencing data as an initial constraint. By extracting salient local features (such as points, lines, or regions) and establishing correspondences through invariant descriptors, they offer enhanced flexibility and resilience against severe geometric distortions.
 
 Unlike region-based approaches, feature-based methods generally do not require georeferencing data as an initial constraint. Instead, they first extract salient features (such as points, lines, or regions) and then establish correspondences by computing descriptors for these features \citep{geng_feature-based_2025}. Consequently, feature detection and description form the core pillars of multimodal matching. This section systematically reviews recent advancements in these two critical steps for remote sensing image registration.
 
 In remote sensing, mainstream feature matching relies on extracting highly repeatable feature points and constructing local invariant descriptors to represent their neighborhoods. Traditionally rooted in computer vision, these classical detection and description techniques depend heavily on image intensity or gradient information. Prominent feature point detectors include Moravec (1980), Harris \citep{harris_combined_1988}, DoG \citep{lowe_distinctive_2004}, and FAST \citep{rosten_faster_2010}. For feature description, SIFT \citep{lowe_distinctive_2004}, GLOH \citep{mikolajczyk_performance_2005}, SURF \citep{bay_speeded-up_2008}, and DAISY \citep{tola_daisy_2010} remain the most widely adopted local invariant descriptors.
 
 To address the unique characteristics of remote sensing imagery, researchers have actively adapted classical feature detection and description operators. \cite{li_sift_2006} pioneered the application of SIFT for remote sensing image registration. Subsequently, \cite{yu_fast_2008} integrated Harris and SIFT detectors within a cross-matching strategy based on wavelet pyramids and triangulated irregular networks (TINs) to register Quickbird, SPOT, and TM images. To improve feature quality, \cite{sedaghat_uniform_2011} proposed Uniform Robust SIFT (UR-SIFT), utilizing stability and significance constraints to extract uniformly distributed features across both scale and image spaces. Building upon this, they later introduced Adaptive Binning SIFT (AB-SIFT) \citep{sedaghat_distinctive_2015}, which employs an adaptive histogram quantization strategy to generate descriptors more robust against local geometric distortions.
 
 Because the aforementioned feature operators rely heavily on image intensity or gradient information, they struggle to achieve high-precision matching on multimodal remote sensing images exhibiting severe geometric distortions and nonlinear radiometric differences. To address these limitations, recent research has introduced a variety of novel multimodal registration methods specifically designed to overcome scale and rotation variations, radiometric discrepancies, and noise interference.
 
 These advanced multimodal registration methods can generally be classified into three categories. The first approach enhances descriptor robustness against nonlinear radiometric differences through gradient optimization \citep{dellinger_sar-sift_2015, xiang_os-sift_2018, yao_multimodal_2022}. The second category leverages local self-similarity (LSS) within log-polar coordinates, extracting internal geometric structures from local neighborhoods to build resilience against radiometric variations \citep{hong_max-index_2022, sedaghat_distinctive_2015, sedaghat_illumination-robust_2019}. Finally, a third group employs PC models to formulate registration algorithms inherently invariant to illumination and contrast fluctuations \citep{yao_heterologous_2021, ye_local_2018, zhang_glift_2025, zhu_r2fd2_2023}.
 
 To summarize these advancements, \autoref{Table:3} provides a comprehensive overview of representative feature-based multimodal registration methods, detailing their core concepts alongside links to available open-source code. The following subsections will review these three methodological categories in detail.

 \begin{table}[htbp]
 	\centering
 	\renewcommand{\arraystretch}{1.3}  
 	\caption{Feature-based registration methods}
 	\label{Table:3}
 	\begin{tabular*}{\textwidth}{@{\extracolsep{\fill}} 
 			>{\centering\arraybackslash}m{1.2cm} 
 			>{\centering\arraybackslash}m{4cm} 
 			>{\centering\arraybackslash}m{8cm} 
 			>{\centering\arraybackslash}m{2cm} @{}}
 		\toprule
 		Method type & Representative algorithm & Core ideas & Acquisition channels \\
 		\midrule
 		\multirow{8}{*}[5ex]{\rotatebox{270}{Gradient optimization}} 
 		& SAR-SIFT \citep{dellinger_sar-sift_2015} & Ratio gradient via ROEWA operator; Robust to SAR speckle noise & \href{https://github.com/ZeLianWen/Image-Registration}{[Code]} \\
 		& PSO-SIFT \citep{ma_remote_2017} & Gradient derivative in Gaussian scale space; Robust to intensity variations & \href{https://github.com/ZeLianWen/Image-Registration}{[Code]} \\
 		& OS-SIFT \citep{xiang_os-sift_2018} & Multi-scale Sobel \& ROEWA operators; Consistent optical-SAR gradients & \href{https://github.com/xym2009/OS-SIFT}{[Code]} \\
 		& CoFSM \citep{yao_multimodal_2022} & Scale space via co-occurrence filtering; Butterworth filter gradient optimization & \href{https://skyearth.org/publication/project/CoFSM/}{[Code]} \\
 		\midrule 
 		\multirow{10}{*}[2ex]{\rotatebox{270}{Local self-similarity}} 
 		& DOBSS \citep{sedaghat_distinctive_2015} & Related value histogram; Internal geometric layout capture & - \\
 		& HOSS \citep{sedaghat_illumination-robust_2019} & Adaptive log-extreme spatial structure \& max correlation rotation index map & \href{https://www.researchgate.net/publication/328733655_HOSS_Descriptor}{[Code]} \\
 		& OSS \citep{xiong_self-similarity_2021} & Self-similar symmetry; Fast calculation via offset mean filtering & - \\
 		& ASS \cite{xiong_robust_2022} & Optimized offset mean filtering; Local statistical weighted differential for speckle suppression & \href{https://gitee.com/xxin08/ASS}{[Code]} \\
 		& MLSS \citep{hong_max-index_2022} & Max value index map \& LSS combination; Radial interval summation & - \\
 		& MOSS \citep{zhangRobustRegistrationMultimodal2024} & Multi-dimensional oriented self-similarity & - \\
 		& ATSC \citep{yang_adjacent_2024} & Adjacent self-similarity 3-D convolution & \href{https://github.com/yangwill81/ASTC}{[Code]} \\
 		\midrule 
 		\multirow{11}{*}[5ex]{\rotatebox{270}{Phase consistency}} 
 		& LHOPC \citep{ye_local_2018} & MMPC-Lap detector \& LHOPC descriptor; Robust to complex radiation differences & - \\
 		& RIFT \citep{li_rift_2020} & Inflection/edge point detection; Maximum Index Map (MIM) construction & \href{https://github.com/LJY-RS/RIFT-multimodal-image-matching}{[Code]} \\
 		& HAPCG \citep{xiong_self-similarity_2021} & Anisotropic weighted moment matrix; Absolute phase coherence direction gradients & \href{https://skyearth.org/publication/project/HAPCG/}{[Code]} \\
 		& MABPC \cite{fan_novel_2022} & Modified Harris-Laplace detector; Multi-scale phase consistency encoding via adaptive spatial structures & - \\
 		& R\textsubscript{2}FD\textsubscript{2} \citep{zhu_r2fd2_2023} & Multi-channel autocorrelation \& log-Gabor; Rotation-invariant RMIM via DAISY & - \\
 		& RIFT2 \citep{liRIFT2SpeedingupRIFT2023} & Replaces convolution sequence ring with a dominant index value & \href{https://github.com/LJY-RS/RIFT-multimodal-image-matching}{[Code]} \\		
 		& GLIFT \citep{zhang_glift_2025} & Log-Gabor filter; FAST; global MIM-based search & \href{https://github.com/wdzsc/GLIFT}{[Code]} \\	
 		& EPC-SRC \citep{ZhengYao2025} & Chunked Harris detector; adaptive weighted moment map; GLOH-like structure & - \\							
 		\bottomrule
 	\end{tabular*}
 \end{table}
 
 \subsection{ Feature registration method based on gradient optimization}
 \cite{dellinger_sar-sift_2015} redefined gradient calculation by replacing traditional differential methods with the ROEWA operator. This adaptation, known as SAR-SIFT, yields gradient magnitudes and orientations that are highly robust to speckle noise, enabling the effective registration of multi-view SAR images. Building on this, \cite{ma_remote_2017} introduced PSO-SIFT (Position Scale Orientation SIFT), which formulates a novel gradient definition within the Gaussian scale space to enhance feature matching. Similarly, \cite{xiang_os-sift_2018} developed OS-SIFT, which leverages multi-scale Sobel and ROEWA operators to generate consistent gradients across optical and SAR imagery. They further constructed a Harris scale space and employed a GLOH-like descriptor to reliably establish correspondences. More recently, \cite{yao_multimodal_2022} proposed CoFSM (Co-occurrence Filter Space Matching), utilizing co-occurrence filters for scale space construction and Butterworth filters for gradient optimization. This effectively mitigates the interference of nonlinear radiometric differences, thereby enhancing multimodal descriptor robustness.
 
 \subsection{Feature registration method based on local self-similarity}
 Focusing on internal geometric structures, \cite{sedaghat_distinctive_2015} substituted gradients with correlation values to generate orientation histograms. By leveraging these correlations, their distinctive order-based self-similarity (DOBSS) descriptor successfully registered multi-sensor satellite imagery. Building upon this concept, \cite{sedaghat_illumination-robust_2019} designed the histogram of oriented self-similarity (HOSS) descriptor. To achieve rotation invariance and enhance matching accuracy, they introduced a Rotation Index of the Maximal Correlation (RIMC) map integrated with an adaptive log-polar spatial structure.
 
 Furthermore, \cite{xiong_self-similarity_2021} proposed the oriented self-similarity (OSS) descriptor, utilizing offset mean filtering to rapidly compute features based on structural symmetry. They subsequently refined this filtering process and employed local statistical weighted differentials to suppress coherent speckle, resulting in the adjacent self-similarity (ASS) descriptor \citep{xiong_robust_2022}. During the same period, \cite{hong_max-index_2022} formulated the max-index based local self-similarity (MLSS) descriptor by combining maximum index mapping with LSS, summing vectors across varying radial intervals at identical angles to achieve highly robust multimodal registration. Expanding upon these oriented feature representations, a novel multi-dimensional oriented self-similarity (MOSS) descriptor was introduced \citep{zhangRobustRegistrationMultimodal2024} to capture richer structural dimensions, thereby further enhancing the robust registration of multi-modal remote sensing images. Most recently, to amplify structural saliency and mitigate noise, \cite{yang_adjacent_2024} integrated 3D Gaussian convolution into the earlier ASS framework. Their resulting ASTC (adjacent self-similarity 3-D convolution) descriptor demonstrated significant experimental gains in robustness, accuracy, and efficiency.

 \subsection{Feature registration method based on phase consistency}
 \cite{ye_local_2018} pioneered the use of the illumination- and contrast-invariant PC model for feature detection and description. Their MMPC-Lap (Minimum Moment of Phase Congruency Lap) detector utilizes auto-scale localization to identify stable keypoints in scale space, while the LHOPC (Local Histogram of Oriented Phase Congruency) descriptor adopts a DAISY-like spatial configuration to maintain robustness under complex radiometric variations. \cite{li_rift_2020} introduced the Radiation-variation Insensitive Feature Transform (RIFT), which detects feature points on PC minimum/maximum moment maps and constructs a Maximum Index Map (MIM) via log-Gabor convolution sequences, significantly improving radiation invariance. To further enhance matching, \cite{yao_heterologous_2021} proposed an anisotropic weighted moment map for detection, coupled with the Histogram of Absolute Phase Congruency Gradients (HAPCG) within a log-polar structure for description.
 
 To mitigate modality-induced detection errors, \cite{fan_novel_2022} integrated nonlinear diffusion and spatial consistency into a Harris-Laplace detector, while their MABPC (Multiscale Adaptive Binning Phase Congruency) descriptor utilizes an adaptive binning structure to encode multiscale features. Addressing the potential loss of multi-directional information in traditional PC moment-based detectors, \cite{zhu_r2fd2_2023} developed the R\textsubscript{2}FD\textsubscript{2} framework. This method employs a multi-channel autocorrelation strategy (MALG) to detect highly repeatable features and introduces a Rotation-invariant Maximum Index Map (RMIM) within a DAISY configuration to achieve radial and rotational invariance. During the same prolific period in 2023, several other significant PC-based frameworks emerged. Notably, to overcome the severe computational bottlenecks of the original RIFT algorithm, \cite{liRIFT2SpeedingupRIFT2023} proposed RIFT2, which replaces the convolution sequence ring with a dominant index value technique, dramatically reducing both memory consumption and processing time. Concurrently, \cite{jiaOpticalSARImage2023} developed the MSPCO descriptor by building a nonlinear diffusion scale space to better preserve edge information, coupled with an effective scale ratio strategy that significantly boosts computational efficiency. To further address large-scale and rotation discrepancies, \cite{xieOpticalSARImage2023} introduced a multi-scale PC-Harris keypoint extractor integrated with a novel polar coordinate-based descriptor. Moving to the latest advancements, \cite{zhang_glift_2025} proposed the GLIFT method, which combines log-Gabor filtering with the FAST algorithm for keypoint detection. By integrating a global MIM-based search with local pixel-level cosine similarity, GLIFT achieves superior results in both match count and precision. In a parallel 2025 advancement, \cite{ZhengYao2025} formulated the Extended Phase Consistency Spatial Relationship Constraints (EPC-SRC) algorithm. By utilizing a chunked Harris detector on an adaptive weighted moment map and constructing a multiscale weighted MIM (MSW-MIM) with a GLOH-like structure, EPC-SRC substantially enhances descriptor distinguishability and matching accuracy.
 
 To synthesize these characteristics and better guide algorithm selection across specific multimodal challenges, the practical utility of these traditional handcrafted descriptors must be weighed against their inherent trade-offs. Gradient-based optimization methods (e.g., SAR-SIFT, OS-SIFT) excel in mitigating specific interference patterns, such as SAR speckle, making them highly effective for targeted, sensor-specific noise reduction; however, this tailored nature often restricts their generalizability across diverse multimodal scenarios. Meanwhile, local self-similarity (LSS) approaches are particularly well-suited for capturing internal geometric layouts, providing reliable resilience against moderate radiometric shifts and structural distortions, yet they frequently struggle under severe nonlinear radiometric variations due to the lower discriminative power of their descriptors. In contrast, Phase Congruency (PC)-based methods demonstrate the highest robustness against severe nonlinear radiometric variations and extreme illumination changes. Although PC methods generally incur a higher computational overhead than gradient or LSS approaches—a limitation actively being mitigated by recent algorithmic innovations like memory-optimized RIFT2—they remain the optimal, most reliable choice and the robust cornerstone for aligning highly heterogeneous multimodal image pairs.
 
 \section{Deep Learning-based Registration Methods}
 While traditional feature-based methods significantly improve robustness against scale and rotation variations compared to region-based techniques, their reliance on handcrafted descriptors limits their generalizability under severe, complex nonlinear radiometric discrepancies. To address these representational bottlenecks, researchers have increasingly turned to deep learning. Driven by rapid advancements in the field, architectures such as Convolutional Neural Networks (CNNs), Fully Convolutional Networks (FCNs), Generative Adversarial Networks (GANs), Siamese Networks (SNs), and Transformers have demonstrated unparalleled capabilities in feature learning. Consequently, these technologies have been extensively integrated into various remote sensing tasks, including photogrammetry, land-cover classification, target recognition, and change detection \citep{heipke_deep_2020,ma_deep_2019,hanHFNetSemanticDifferential2025}. By leveraging these powerful, data-driven neural architectures to extract high-level semantic representations, deep learning approaches can more effectively bridge the severe heterogeneous gaps and mitigate the complex noise that traditional handcrafted features often struggle to capture.
 
 By adopting a learning-feedback-adjustment paradigm, data-driven deep learning effectively extracts high-level semantic information from vast datasets, gaining significant traction in multimodal remote sensing registration \citep{zhu_advances_2023}. Researchers leverage these deep architectures to extract more refined common features than traditional handcrafted (i.e., shallow) descriptors, thereby better mitigating the discrepancies caused by complex geometric distortions, radiometric variations, and noise.
 
 In this paper, we categorize deep learning-based registration into two frameworks: modular methods and end-to-end methods. Modular approaches integrate deep learning into specific stages of the traditional registration pipeline—replacing individual components within region- or feature-based workflows (e.g., using CNNs, GANs, or Siamese networks for feature extraction or matching). In contrast, end-to-end methods circumvent staged processing by constructing unified neural network architectures that directly output registration results from raw input images.
 
 \subsection{Modular registration methods}
 The most prevalent modular strategy involves embedding deep neural networks into traditional feature- or region-based pipelines. By leveraging the data-driven nature of deep learning, these methods extract high-dimensional latent features to yield more robust feature detectors, descriptors, or similarity metrics. This modular integration effectively capitalizes on the superior representation capabilities of deep learning to enhance the overall robustness of the registration process.
 
 \begin{table}[htbp]
 	\centering
 	\renewcommand{\arraystretch}{1.3}  
 	\caption{Learning-based template matching}
 	\label{Table:DLtemplate}
 	\begin{tabular*}{\textwidth}{@{\extracolsep{\fill}}  
 			>{\centering\arraybackslash}m{3.5cm} 
 			>{\centering\arraybackslash}m{10cm} 
 			>{\centering\arraybackslash}m{2cm} @{}}
 		
 		\toprule
 		Representative algorithm & Core ideas & Acquisition channels\\
 		\midrule 
 		\cite{merkle_exploiting_2017} & Siamese FCN; Dot product similarity for optical-SAR offsets & - \\	
 		\cite{hughes_identifying_2018} & Independent convolutional streams (Pseudo-Siamese); Accommodates image heterogeneity & - \\	
 		SFCNet \citep{zhang_registration_2019} & Siamese FCN with convolutional connections; Novel feature distance loss & - \\	
 		\cite{li_deep_2021} & Semantic template matching framework & \href{https://github.com/liliangzhi110/semantictemplatematching}{[code]} \\
 		MSF SiamUNet-7 \citep{liu_swin_2021} & MSF SiamUNet-7; Phase-structure features; Local texture \& global structure fusion & - \\ 
 		OSMNet \citep{zhang_explore_2022} & Multi-frequency channel attention; adaptive weighted loss; Structural consistency & \href{https://github.com/zhanghan9718/OSMNet}{[Code]} \\	
 		DDFN \citep{zhang_optical_2022} & Siamese-based deep dense feature network; Deep feature extraction & - \\	
 		MCGF \citep{zhou_robust_2022} & Pseudo-Siamese network; Multi-scale convolutional gradient features; FFT acceleration & - \\
 		PCNet \citep{cao_pcnet_2023} & Siamese network; Enhances structural similarity; SSD matching & - \\	
 		SIFNet \citep{liu_sifnet_2023} & ResNet-FPN; feature pyramids; encoding-decoding process & - \\	
 		\cite{yangAccurate2024}  & Reframing template matching process as a spatial alignment task & - \\	
 		\cite{wang_multihierarchy_2025}  & Combined ResNet and FPN with  Swin-Transformer & - \\	
 		
 		\bottomrule 
 	\end{tabular*}
 \end{table}

 \paragraph{Learning-based Template Matching Methods \\}
 Most learning-based template matching methods utilize (pseudo-)Siamese neural architectures \citep{liu_optical_2023, merkle_exploiting_2017, wang_multihierarchy_2025}. The core principle involves training these networks to extract deep latent features that replace traditional intensity or shallow structural information, followed by assessing similarity between the resulting feature maps. This assessment is typically executed via neural layers, dot products, convolutions, or classical similarity metrics. Early approaches, such as \cite{merkle_exploiting_2017}, employed a Siamese fully convolutional network (FCN) with dilated convolutions to enlarge the receptive field, computing similarity via dot products. To address the inherent heterogeneity between SAR and optical imagery, \cite{hughes_identifying_2018} implemented a pseudo-Siamese architecture—consisting of two independent but identical streams—rather than a weight-sharing Siamese network. To mitigate spatial information loss caused by pooling during feature extraction, \cite{nanTemplateMatchingMethod2019} adopted fully convolutional architectures to maintain feature maps at their original input resolution. Similarly, \cite{zhang_registration_2019} utilized SFCNet (Siamese Fully Convolutional Network), employing convolutional layers to effectively correlate feature pairs.
 
 As deep learning techniques matured, the focus shifted toward richer semantic representations and enhanced localization precision. \cite{li_deep_2021} introduced a semantic template matching framework that explicitly leverages high-level semantic features to robustly handle radiometric differences in remote sensing image registration. Further advancing the balance between high-level semantic abstraction and low-level detail preservation, \cite{zhang_optical_2022} developed the Deep Dense Feature Network (DDFN). Concurrently, \cite{zhou_robust_2022} introduced Multiscale Convolutional Gradient Features (MCGFs) within a fully convolutional pseudo-Siamese framework to better accommodate nonlinear radiometric variations. Building on multi-level feature extraction, \cite{zhang_explore_2022} proposed OSMNet, integrating multilayer feature fusion and multi-frequency channel attention with an adaptive weighted loss. Focusing on structural consistency, \cite{cao_pcnet_2023} developed PCNet to enhance the structural similarity of multimodal images before employing the sum of squared differences (SSD) for registration. Alongside this, \cite{liu_optical_2023} combined log-Gabor filters with a multiscale fusion network (MSF SiamUNet-7) to integrate multi-directional phase information. More recently, to fundamentally refine matching accuracy, \cite{yangAccurate2024} proposed a novel method based on center-point localization, effectively reframing the template matching process as a highly robust and precise spatial alignment task.
 
 The emergence of Transformers has further expanded the multimodal registration toolkit \citep{vaswani_attention_2017}. For instance, SIFNet \citep{liu_sifnet_2023} utilizes a ResNet-FPN backbone to construct feature pyramids, incorporating positional encoding and a two-stage encoding-decoding process to facilitate cross-modal interaction. This approach demonstrates substantial performance gains over traditional methods. Extending this paradigm, \cite{wang_multihierarchy_2025} combined ResNet and FPN with the computationally efficient Swin-Transformer attention mechanism \citep{liu_swin_2021} to fuse multiscale features. By incorporating a Spatial Transformer Network (STN) for resampling, their Multihierarchy Flow Field Prediction Network (MFFPN) achieved highly competitive results.
 
 Despite their success in overcoming nonlinear radiometric differences through powerful deep representations, these learning-based template matching methods remain constrained by their limited adaptability to significant scale and rotation variations between multimodal images.
 
 \begin{table}[htbp]
 	\centering
 	\renewcommand{\arraystretch}{1.3}  
 	\caption{Learning-based feature matching}
 	\label{Table:DLFeature}
 	\begin{tabular*}{\textwidth}{@{\extracolsep{\fill}}  
 			>{\centering\arraybackslash}m{5cm} 
 			>{\centering\arraybackslash}m{8.5cm} 
 			>{\centering\arraybackslash}m{2cm} @{}}
 		\toprule
 		Representative algorithm & Core ideas & Acquisition channels \\
 		\midrule
 		\cite{ye_remote_2018} & CNN features (FC \& conv layers); SIFT & - \\	
 		\cite{dong_local_2019} & DescNet deep feature vectors; Robust keypoint descriptors & - \\
 		\cite{maNovelTwoStepRegistration2019a} & Deep \& handcrafted feature combination; Fine matching \& transformation & - \\	
 		CMM-Net \citep{lan_deep_2021} & CNN high-dimensional feature maps; Keypoint detection \& description & \href{https://github.com/lan-cz/cnn-matching}{[Code]} \\	
 		MAP-Net \citep{cui_map-net_2022} & SPAP module context integration; Attention-weighted dense features & - \\
 		Sim-CSPNet \citep{xiang_algorithm_2022} & Sim-CSPNet architecture; Rapid deep \& shallow feature extraction & - \\	
 		M2DT-Net \citep{zhang_multimodal_2022} & Fine-tuned CNN invariant features; DEGENSAC mismatch removal & - \\
 		Cnet \citep{quan_novel_2023} & Attention learning module; Enhanced feature representation & - \\
 		MatcherTF \citep{liang_coarse--fine_2023} & Transformer (self \& cross-attention); Local feature matching network & - \\
 		\cite{li_radiation-variation_2024} & ResNet-Transformer architecture & \href{https://github.com/JasonLi-UCAS/NIR-VIS-RS}{[Code]} \\
 		SFA-Net \citep{gao_sfa-net_2025} & SuperPoint; Segment Anything Model & \href{https://github.com/GaotTian/SFA-Net}{[Code]} \\
 		MINIMA \citep{renMINIMA2025} & Shifting the paradigm from architectural complexity to data-driven generalization & \href{https://github.com/LSXI7/MINIMA}{[Code]} \\
 		\bottomrule 
 	\end{tabular*}
 \end{table}
 
 \paragraph{Learning-based Feature Matching Methods \\}
 Learning-based feature matching methods primarily leverage deep architectures to generate trainable detectors or descriptors, significantly enhancing registration performance \citep{geng_feature-based_2025}.
 
 Early explorations by \cite{ye_remote_2018} analyzed fully connected and convolutional layer features across various aggregation scales, integrating them with SIFT to achieve successful multi-sensor registration. Similarly, \cite{dong_local_2019} utilized the DescNet architecture to generate high-dimensional feature vectors. By replacing traditional handcrafted descriptors with these more discriminative deep vectors, they significantly increased the number of correct correspondences. To bridge global relationships and local precision, \cite{maNovelTwoStepRegistration2019a} proposed a coarse-to-fine framework. It utilizes a VGG-16 backbone for pyramidal feature detection to approximate spatial relationships, followed by a combination of deep and handcrafted features for refined transformation estimation.
 
 Advancing beyond simple feature replacement, \cite{lan_deep_2021} developed CMM-Net (Cross-Modality Matching Net), drawing inspiration from the D2-Net ``detect-and-describe" paradigm to extract keypoints and descriptors simultaneously from high-dimensional CNN feature maps. \cite{cui_map-net_2022} introduced MAP-Net, which incorporates Spatial Pyramid Aggregated Pooling (SPAP) and an attention mechanism. The SPAP module integrates global and local contexts, while the attention mechanism weights dense features to extract advanced semantic representations that are highly adaptable to geometric and radiometric variations.
 
 Efficiency and precision have also been key drivers of innovation. \cite{xiang_algorithm_2022} designed Sim-CSPNet (Siamese Cross Stage Partial Network) to rapidly extract hybrid deep-shallow descriptors for multi-resolution SAR images. \cite{zhang_multimodal_2022} developed M2DT-Net, which utilizes a fine-tuned CNN for robust invariant feature learning, coupled with adaptive distance constraints and DEGENSAC refinement to filter outliers. Addressing correlation learning, \cite{quan_deep_2022} proposed Cnet, utilizing an attention-based convolutional network to focus on semantically meaningful features. \cite{quan_novel_2023} further extended this into a coarse-to-fine framework featuring rotation rectification based on deep ordinal regression (DOR) and fine registration via Deep Descriptor Learning (DDL). By treating rotation estimation as an ordinal regression problem, the DOR network effectively improves the accuracy of initial orientation alignment.
 
 The integration of Transformers has recently revolutionized feature aggregation through their global attention mechanisms. \cite{liang_coarse--fine_2023} developed MatcherTF, which employs self- and cross-attention to generate highly discriminative keypoint descriptors. Similarly, \cite{zhang_multimodal_2024} proposed Modality-Independent Consistency Matching (MICM), utilizing a U-Net backbone with Transformer-based encoding-decoding for feature aggregation. Following this, \cite{li_radiation-variation_2024} introduced a robust ResNet-Transformer architecture. Exploring further hybridizations, CTNet \citep{yue_ctnet_2025} successfully fuses CNN-based local information with Transformer-based global contexts. Beyond purely structural innovations, recent advancements have also embedded physical constraints into Transformer architectures. For instance, combining geographic location attention with phase consistency has proven highly effective in significantly improving generalization and matching accuracy for complex multi-source SAR imagery \citep{gaoArcticSeaIce2025}. Shifting the paradigm from this architectural complexity toward data-driven generalization, \cite{renMINIMA2025} recently introduced MINIMA (Modality Invariant Image Matching). Rather than designing modality-specific modules, MINIMA leverages a generative data engine to transform easily obtainable RGB data into a comprehensive synthetic dataset (MD-syn) encompassing diverse modalities. By directly training advanced pipelines on these randomly selected modality pairs, their framework achieves exceptional zero-shot cross-modal generalization, successfully bridging severe semantic gaps across various heterogeneous scenarios. Ultimately, these studies collectively highlight the evolving landscape of multimodal registration, progressing from complex attention mechanisms to highly scalable, data-centric learning paradigms.
 
 Self-supervised frameworks like SuperPoint \citep{detoneSuperPointSelfSupervisedInterest2018} have also been adapted for remote sensing. To address SuperPoint's inherent limitations in cross-modal representation, \cite{gao_sfa-net_2025} developed SFA-Net, which augments SuperPoint features with edge structures extracted by the Segment Anything Model (SAM). By processing these through a regional self-attention network, they achieved superior performance across multiple metrics.
 
 In summary, learning-based feature matching significantly improves detector repeatability and descriptor robustness by leveraging high-dimensional semantic representations. However, it is noteworthy that current research in multimodal registration remains predominantly focused on feature description, with relatively less exploration into specialized feature detection for heterogeneous data.
 
 \paragraph{Modality Unification Methods Based on Style Transfer \\}
 To mitigate the impact of significant nonlinear radiometric differences, generative models such as Generative Adversarial Networks (GANs) have been introduced to the field. These approaches achieve modality unification through style transfer, subsequently employing traditional region- or feature-based techniques to register the transformed images. Specifically, these methods leverage image-to-image translation to ensure that the multimodal images exhibit consistent radiometric characteristics and structural features. For instance, \citet{merkle_exploring_2018} utilized Conditional GANs (CGANs) to generate pseudo-SAR image patches from optical data, achieving superior matching accuracy using Normalized Cross-Correlation (NCC) and SIFT features. Similarly, \citet{zhang_multimodal_2019} extended visual attribute transfer by fusing original structure and texture info, thereby weakening nonlinear radiometric discrepancies and effectively improving the precision of traditional feature-based matching.
 
 \begin{table}[htbp]
 	\centering
 	\renewcommand{\arraystretch}{1.3}  
 	\caption{Modality Unification Methods Based on Style Transfer}
 	\label{Table:StyleTransfer}
 	\begin{tabular*}{\textwidth}{@{\extracolsep{\fill}}  
 			>{\centering\arraybackslash}m{4.5cm} 
 			>{\centering\arraybackslash}m{11cm} @{}}
 		\toprule
 		Representative algorithm & Core ideas \\
 		\midrule
 		\cite{ararUnsupervisedMultiModalImage2020} & Special loss function; Strict geometric position preservation; Texture style change \\		
 		\cite{zhao_intelligent_2022} & Smooth CycleGAN; Novel loss function; Improved generation quality \\
 		\cite{chenVisibleInfraredImage2022} & Modified CycleGAN; Identity mapping loss for color constraint \\
 		CDA-GAN \citep{wuOpticaltoSARTranslationBased2024} & Attention mechanism; High-frequency texture retention; Pseudo-SAR generation \\
 		CSTNet \citep{bi_unsupervised_2025} & Visible-to-pseudo-infrared conversion; Unified modal features; Multi-scale cascaded network \\
 		Seg-CycleGAN \citep{zhangSegCycleGANSARtoOpticalImage2025} & Semantic segmentation guidance; SAR-to-optical conversion; Structural information retention \\
 		\bottomrule 
 	\end{tabular*}
 \end{table}
 
 \citet{zhao_intelligent_2022} proposed a Smooth Cycle-Consistent GAN with a refined loss function to improve the quality of synthesized images. By employing D2-Net and LoFTR for matching between style-transferred and original aerial images, they effectively enhanced registration accuracy and robustness. Similarly, \citet{chenVisibleInfraredImage2022} developed a GAN-based matching method for visible and infrared imagery, reducing cross-modal matching difficulty through a custom loss function. \citet{bi_unsupervised_2025} introduced the Cross-modal Style Transfer Network (CSTNet), which utilizes a U-Net generator for optical-to-infrared translation. Their approach ensures structural consistency and modal authenticity via a multi-scale discriminator, while incorporating $L_2$ loss to constrain pixel-level similarity. More recently, \citet{wei_osdm-mreg_2025} designed an unaligned Target-Guided One-Step Condition Diffusion Model (UTGOS-CDM) to achieve modality translation.
 
 Operationally, the typical workflow of style transfer-based methods relies on a decoupled, two-step pipeline. First, a computationally intensive generative model—such as a GAN or a Diffusion model—is employed to translate one modality into the other (e.g., converting an optical image into a pseudo-SAR image), thereby unifying their radiometric characteristics. Second, mature traditional matching algorithms (e.g., SIFT, CFOG, or NCC) are applied directly to the unified image pair to extract features, establish correspondences, and estimate the final geometric transformation.
 
 While this modular design offers the technical advantage of plug-and-play flexibility with highly interpretable, established matching algorithms, it inherently incurs substantial computational overhead. Qualitatively, when compared to direct end-to-end methods—which execute feature extraction and spatial alignment simultaneously within a single, highly optimized neural forward pass—style transfer approaches necessitate the sequential execution of a heavy generative network followed by an iterative traditional matching search. Consequently, although this strategy effectively mitigates nonlinear radiometric differences, its cumulative computational cost and high latency remain significant limitations for real-time or resource-constrained applications.
 
 Beyond these computational constraints, the efficacy of style transfer-based modality unification remains heavily reliant on the scale and diversity of its training data. Insufficient sample sizes or limited scene coverage frequently lead to degraded modality conversion quality. This inconsistent translation performance across varying geographic scenes and sensor types persists as a critical bottleneck, ultimately compromising the robustness and precision of the subsequent registration process \citep{velesaca_multimodal_2024}. To circumvent the vulnerabilities inherent in this decoupled paradigm, contemporary research is increasingly shifting away from pure style translation toward unified, end-to-end registration frameworks. By jointly optimizing cross-modal feature extraction and spatial alignment within a single architecture, these modern approaches bypass the need for intermediate image generation, thereby significantly enhancing both generalizability and overall matching efficiency.
 
 \subsection{End-to-end registration methods}
 Increasingly, research has pivoted toward direct estimation of geometric transformation parameters or deformation fields, a trend primarily rooted in computer vision and medical imaging \citep{ma_image_2021}. Recent studies have extended this paradigm to remote sensing \citep{hughes_deep_2020,li_multimodal_2023,wang_deep_2018}, with representative end-to-end heterogeneous registration methods summarized in \autoref{Table:end2end}.
 
 \citet{wang_deep_2018} proposed an end-to-end model that unifies feature extraction and matching within a single network. By establishing a closed loop of information feedforward and feedback, the model directly learns the mapping between image pairs and their corresponding matching labels. To overcome the scarcity of remote sensing training data, a self-learning strategy was introduced, while transfer learning was employed to mitigate the computational burden during training. Similarly, \citet{li_remote_2022} proposed D-DenseNet, a weight-sharing network designed for robust multimodal feature extraction. By pairing reference and target image patches, the network directly regresses the displacement parameters of the four corner points, enabling fully end-to-end training. Taking a sequential approach, \citet{hughes_deep_2020} introduced a framework for sparse SAR-optical matching comprising three specialized networks: the first identifies salient feature regions; the second performs multi-scale matching based on a feature-space correlation function; and the third functions as an error-rejection module to evaluate matching quality and eliminate significant outliers.
 
 \begin{table}[htbp]
 	\centering
 	\renewcommand{\arraystretch}{1.3}  
 	\caption{End-to-end registration methods}
 	\label{Table:end2end}
 	\begin{tabular*}{\textwidth}{@{\extracolsep{\fill}}  
 			>{\centering\arraybackslash}m{4.5cm} 
 			>{\centering\arraybackslash}m{9cm} 
 			>{\centering\arraybackslash}m{2cm} @{}}
 		\toprule
 		Representative algorithm & Core ideas & Acquisition channels \\
 		\midrule 
 		\cite{wang_deep_2018} & Closed-loop feedforward/feedback mapping; Self-learning for small datasets & - \\
 		\cite{hughes_deep_2020} & Sequential networks: feature extraction, multi-scale matching, error rejection & - \\
 		D-DenseNet \citep{li_remote_2022} & Weight-sharing D-DenseNet; Direct learning of 4-corner displacement parameters & - \\
 		\cite{li_deep_2021} & Semantic template matching framework; Spatial location probability mapping & \href{https://github.com/liliangzhi110/semantictemplatematching}{[Code]} \\
 		MU-Net \citep{ye_multiscale_2022} & Unsupervised MU-Net; Multi-scale parameter learning; Structural similarity loss & \href{https://github.com/yeyuanxin110/MU-Net}{[Code]} \\	
 		\cite{li_multimodal_2023} & Self-attention fusion framework; Cross-modal feature regression via point displacements & \href{https://github.com/liliangzhi110/E2EIR}{[Code]} \\	
 		\cite{zheng_plug-and-play_2025} & plug-and-play; U-Net structure; modality prompt module and detail enhancement module & \href{https://github.com/stwts/PGMR}{[Code]} \\
 		GDROS\citep{sunGDROS2025} & CNN-Transformer hybrid module; least squares regression module & \href{https://github.com/Zi-Xuan-Sun/GDROS}{[Code]} \\		
 		PGMR\citep{zheng_plug-and-play_2025} & Modality Prompt Module; Detail Enhancement Module & \href{https://github.com/stwts/PGMR}{[Code]} \\		
 		\bottomrule 
 	\end{tabular*}
 \end{table}
 
 \cite{li_deep_2021} developed a deep learning-based semantic template matching framework that bypasses traditional feature extraction, matching optimization, and pixel-wise similarity searches. By directly mapping the spatial probability of templates onto the reference image, the framework estimates transformation parameters with high precision and computational efficiency. Similarly, \cite{ye_multiscale_2022} introduced MU-Net, an unsupervised, multi-scale network that employs a coarse-to-fine registration strategy. By stacking multiple deep neural network (DNN) models, MU-Net prevents backpropagation from converging to local optima, while its structural similarity-based loss function effectively addresses geometric and radiometric discrepancies. Building on this momentum, \cite{li_multimodal_2023} proposed a fusion-based matching network that leverages a self-attention mechanism to extract cross-modal feature representations. In this approach, registration is formulated as a regression of specified point displacements, enabling an end-to-end corrective process within the network.
 
 Moving into the latest advancements of 2025, researchers have increasingly focused on tackling severe geometric distortions and detail preservation in highly heterogeneous data. To address the critical challenge of large geometric transformations in optical-SAR pairs, \cite{sunGDROS2025} proposed GDROS, a geometry-guided dense registration framework. By utilizing a CNN-Transformer hybrid module, they construct and iteratively refine a multi-scale 4D correlation volume to establish pixel-wise dense correspondences. Furthermore, they implemented a least squares regression (LSR) module to impose geometric constraints on the predicted dense optical flow, effectively mitigating prediction divergence. Concurrently, \cite{zheng_plug-and-play_2025} introduced PGMR, a plug-and-play multimodal registration method. Built on a ``3-layer shared encoder and 3-layer deformation decoder" architecture, PGMR utilizes a Modality Prompt Module (MPM) for feature interaction via a U-Net structure, followed by a Detail Enhancement Module (DEM) to recover fine-grained information lost during resampling.
 
 In summary, while modular registration methods offer flexibility and ease of training—allowing for the integration of specialized modules to meet specific requirements—they are susceptible to error accumulation across multiple stages, often leading to local optima. In contrast, end-to-end registration methods consolidate the entire matching process into a unified neural architecture. Although continuous information feedback during training facilitates global optimization and mitigates local optima, these methods are often constrained by high computational costs, complex architectures, and limited interpretability.
 
 \section{Multimodal image registration datasets}
 This section summarizes several widely used open-source datasets for remote sensing image registration, with detailed specifications and download links provided in \autoref{Table:dataset}. These datasets vary significantly in scale and purpose: large-scale repositories, containing tens of thousands of multimodal image pairs, are ideal for training deep learning models across various tasks, including registration, fusion, and object recognition. In contrast, smaller-scale datasets—comprising dozens to hundreds of pairs across diverse domains and coverage scenarios—serve as essential benchmarks for testing and evaluating algorithmic performance.
 
 \begin{table}[tbp]
 	\centering
 	\renewcommand{\arraystretch}{1.3}  
 	\caption{Multimodal image registration dataset}
 	\label{Table:dataset}
 	\begin{tabular*}{\textwidth}{@{\extracolsep{\fill}} 
 			>{\centering\arraybackslash}m{3.5cm} 
 			>{\centering\arraybackslash}m{2.2cm} 
 			>{\centering\arraybackslash}m{2.2cm}  
 			>{\centering\arraybackslash}m{1.8cm}  
 			>{\centering\arraybackslash}m{2.2cm}  
 			>{\centering\arraybackslash}m{2.5cm} @{}} 
 		\toprule
 		Datasets & Image types & Quantity & Size & Resolution & Download \\
 		\midrule
 		SARptical \citep{wang_sarptical_2018} & Optical-SAR & 10,108 pairs & 112$\times$112 & Opt 0.2m / SAR 1m & \href{https://syncandshare.lrz.de/getlink/figixjRV9idETzPgG689dGB/SARptical_data.zip}{[Link]} \\
 		SEN1-2 \citep{schmitt_sen1-2_2018} & Optical-SAR & 282,384 pairs & 256$\times$256 & 10m & \href{https://mediatum.ub.tum.de/1436631}{[Link]} \\
 		University-1652 \citep{zheng_university-1652_2020} & Multiple & 1,652 buildings & - & - & \href{https://github.com/layumi/University1652-Baseline}{[Link]} \\
 		OS-dataset \citep{xiang_automatic_2020} & Optical-SAR & 10,692 pairs & 256$\times$256 & 1m & \href{https://pan.baidu.com/share/init?surl=4bqaJhMSZEy7EXcXVAc77w}{[Link(vriw)]} \\
 		QXS-SAROPT \citep{huang_qxs-saropt_2021} & Optical-SAR & 20,000 pairs & 256$\times$256 & 1m & \href{https://github.com/yaoxu008/QXS-SAROPT}{[Link]} \\
 		Multimodal Image Matching Datasets \citep{jiang_review_2021} & Multiple & 164 pairs & - & - & \href{https://github.com/StaRainJ/multimodality-image-matching-database-metrics-methods/blob/master/Multimodal_Image_Matching_Datasets/}{[Link]} \\
 		SOPatch \citep{xu_sar-optical_2023} & Optical-SAR & $>$650,000 pairs & - & - & \href{https://pan.baidu.com/share/init?surl=2yIheGOg6JTTsYAQfX7Pfg&pwd=1234}{[Link(1234)]} \\
 		OSEval \citep{xiang_global--local_2023}  & Optical-SAR & Hundreds of pairs & - & Sub-meter & \href{https://github.com/xym2009/OS-Eval}{[Link]} \\
 		3MOS \citep{ye3MOSMultisourceMultiresolution2025} & Optical-SAR & 113,000 pairs & 256$\times$256 & $3.5m\sim12.5m$ & 
 		\href{https://github.com/3M-OS/3MOS}{[Link]} \\
 		\bottomrule
 	\end{tabular*}
 \end{table}
 
\begin{figure}[htbp]
 	\centering
 	\begin{minipage}[t]{0.48\textwidth}
 		\centering
 		\includegraphics[width=\textwidth]{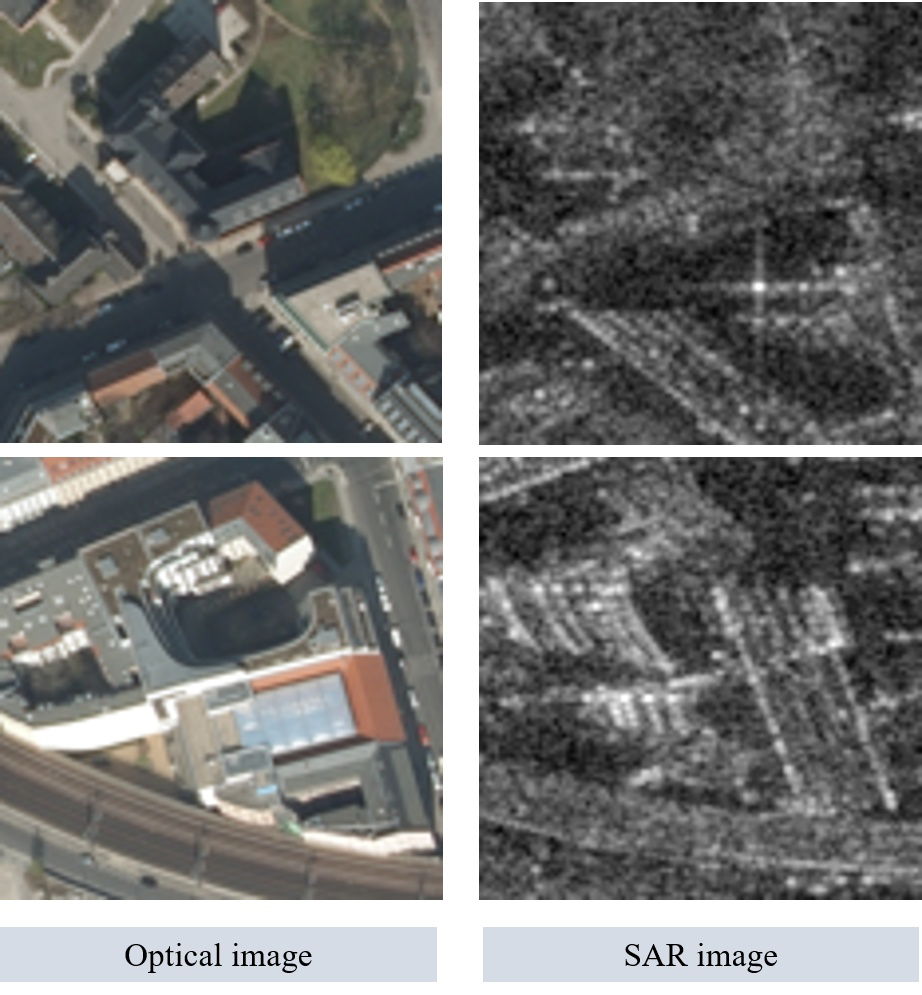}
 		\caption{Examples of SARptical dataset}
 		\label{Fig:sarptical}
 	\end{minipage}
 	\hfill
 	\begin{minipage}[t]{0.48\textwidth}
 		\centering
 		\includegraphics[width=\textwidth]{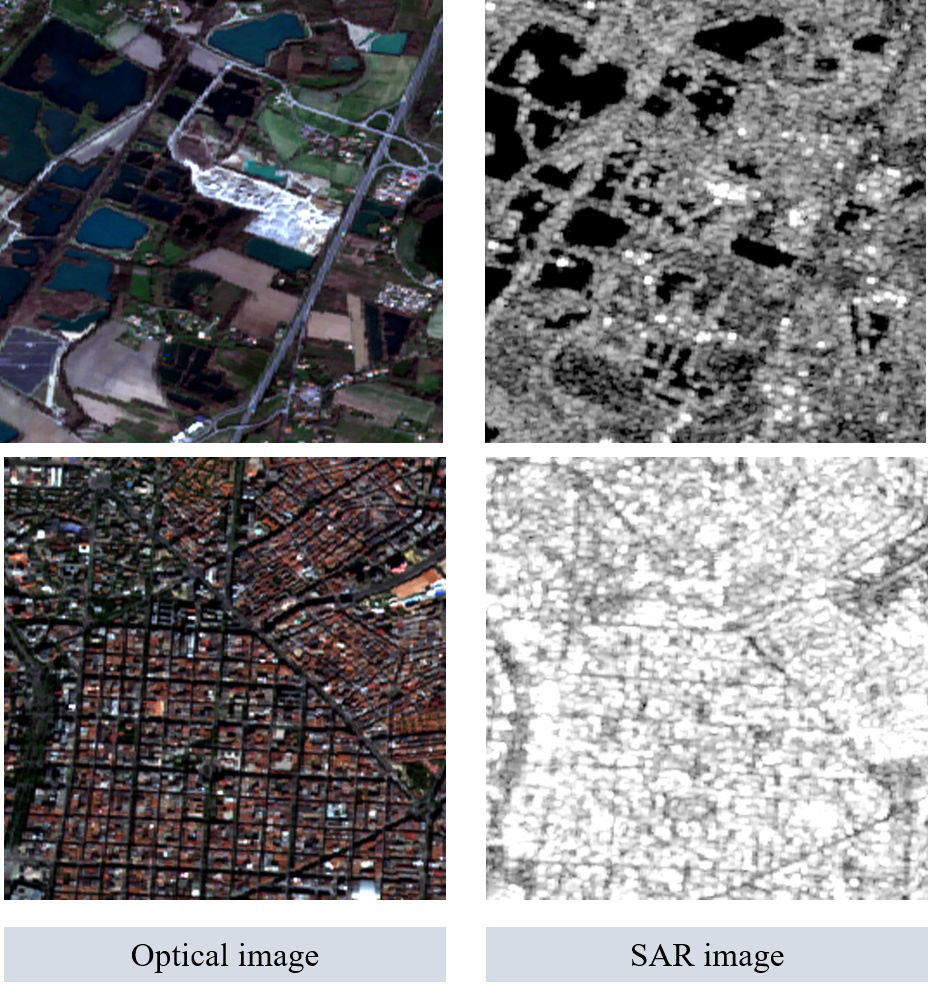}
 		\caption{Examples of SEN1-2 dataset}
 		\label{Fig:sen1-2}
 	\end{minipage}
 \end{figure}

 The SARptical dataset \citep{wang_sarptical_2018} provides a collection of 10,108 strictly aligned optical-SAR image patches. The source data was captured over Berlin, Germany between 2009 and 2013, combining imagery from the TerraSAR-X satellite with airborne UltraCAM photography. The spatial resolution of optical and SAR images are 0.2 meters and 1 meter, respectively, and the pixel size are both 112*112 pixels. \autoref{Fig:sarptical} showcases images of partial scenes in the SARptical data set.
 

 SEN1-2 \citep{schmitt_sen1-2_2018} offers a large-scale benchmark of 282,384 pairs of SAR and optical images taken by the Sentinel series satellites in 2017, covering different surface territory of the five continents in four seasons. The resolution of optical and SAR image is 10 meters, and the picture element size is 256*256 pixels. \autoref{Fig:sen1-2} showcases images of partial scenes in the SEN1-2 dataset.
 
 
 University-1652 \citep{zheng_university-1652_2020} includes data from three platforms: ground images, drone perspective images generated through Google Earth simulation, and satellite images, collected from 1652 buildings across 72 universities worldwide. Each building in this data set has an average of 1 satellite image, 1 drone image, and 3.38 real street view images. It features rich data sources and imaging perspectives, along with clear annotation information. Figure 6 showcases images of partial scenes from this dataset.
 
 \begin{figure}[h]
 	\centerline{\includegraphics[width=1\linewidth]{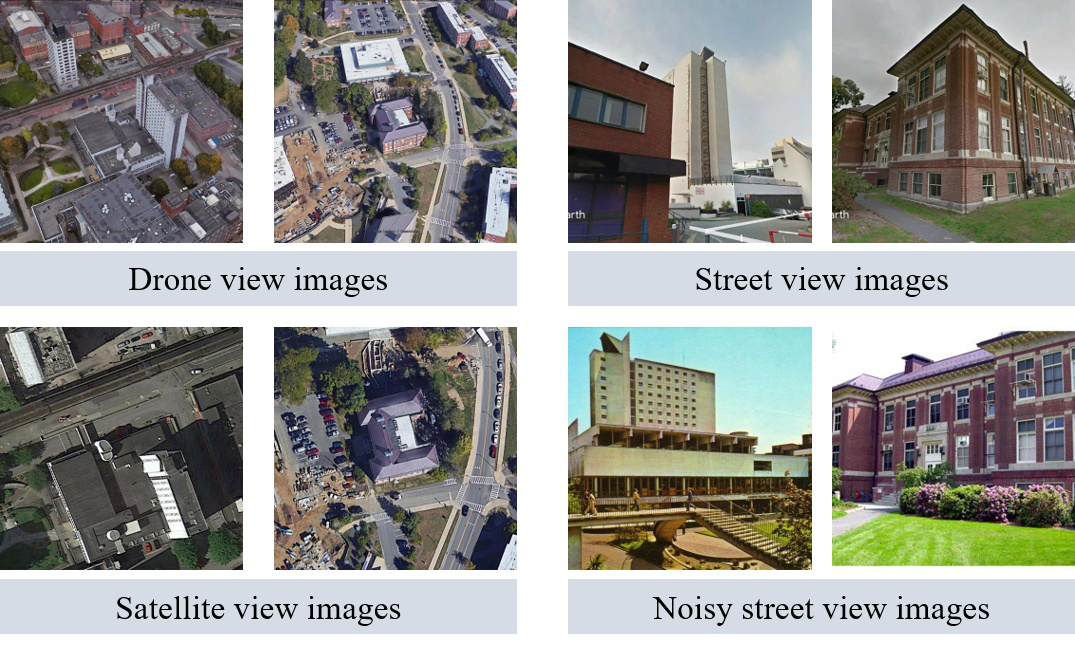}}
 	\caption{Examples of University-1652 dataset}
 	\label{Fig:university1652}
 \end{figure}
 
\begin{figure}[htbp]
 	\centering
 	\begin{minipage}[t]{0.48\textwidth}
 		\centering
 		\includegraphics[width=\textwidth]{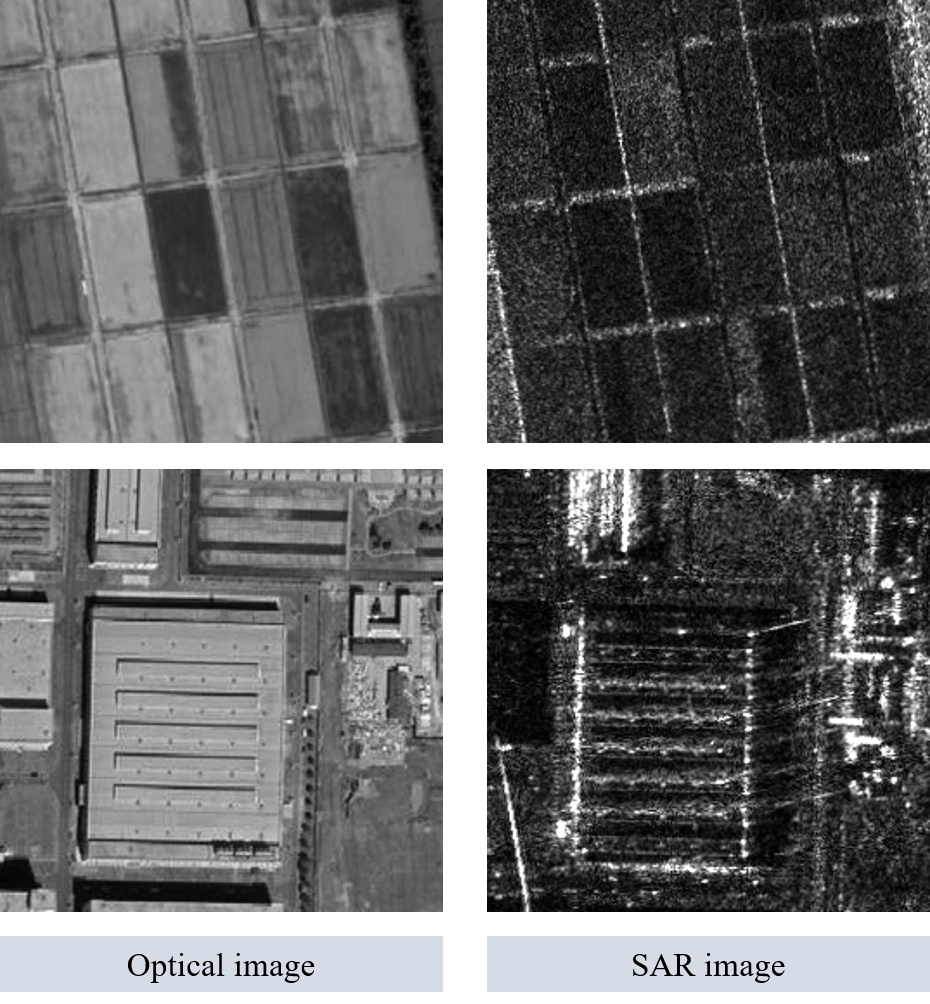}
 		\caption{Examples of OS-dataset}
 		\label{Fig:osdata}
 	\end{minipage}
 	\hfill
 	\begin{minipage}[t]{0.48\textwidth}
 		\centering
 		\includegraphics[width=\textwidth]{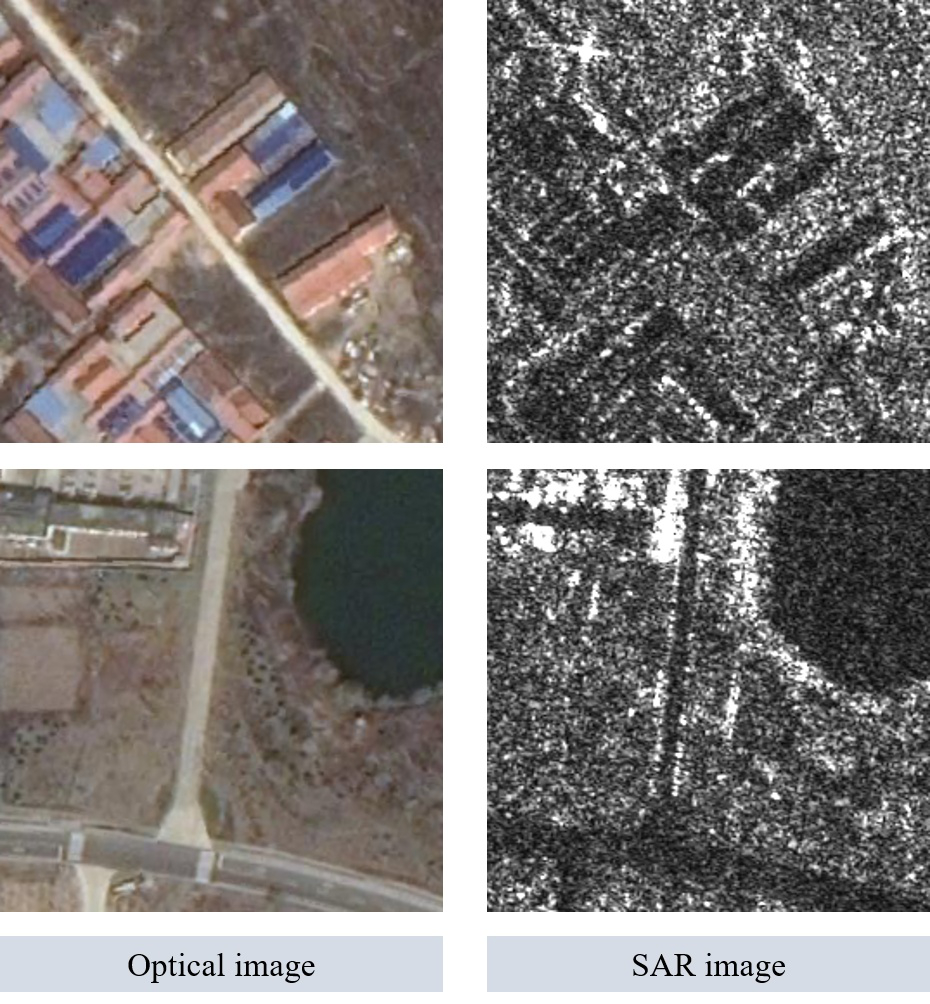}
 		\caption{Examples of QXS-SAROPT dataset}
 		\label{Fig:QXS-SAROPT}
 	\end{minipage}
 \end{figure}

 OS-dataset \citep{xiang_automatic_2020} is a matched dataset consisting of 10,692 pairs of 256*256 pixels hi-res optical and SAR images. Image coverage includes multiple cities worldwide and their surrounding areas. SAR images are from the Gaofen-3 satellite, with a spatial resolution of 1 m optical images were acquired from the Google Earth platform and resampled to 1 m resolution.\autoref{Fig:osdata} showcases images of partial scenes from the OS-dataset. 
 
 
 The QXS-SAROPT \citep{huang_qxs-saropt_2021} dataset consists of 20,000 pairs of 256*256 hi-res optical and SAR images. The QXS-SAROPT data set acquired SAR images from the Gaofen-3 SAR satellite and optical images from Google Earth, maintaining a consistent 1-meter ground resolution for both modalities. Geographically, the data collection focuses exclusively on maritime environments, specifically capturing the port areas of Shanghai, Qingdao, and San Diego. \autoref{Fig:QXS-SAROPT} showcases images of partial scenes from the QXS-SAROPT data set.
 
\begin{figure}[h]
 	\centerline{\includegraphics[width=1\linewidth]{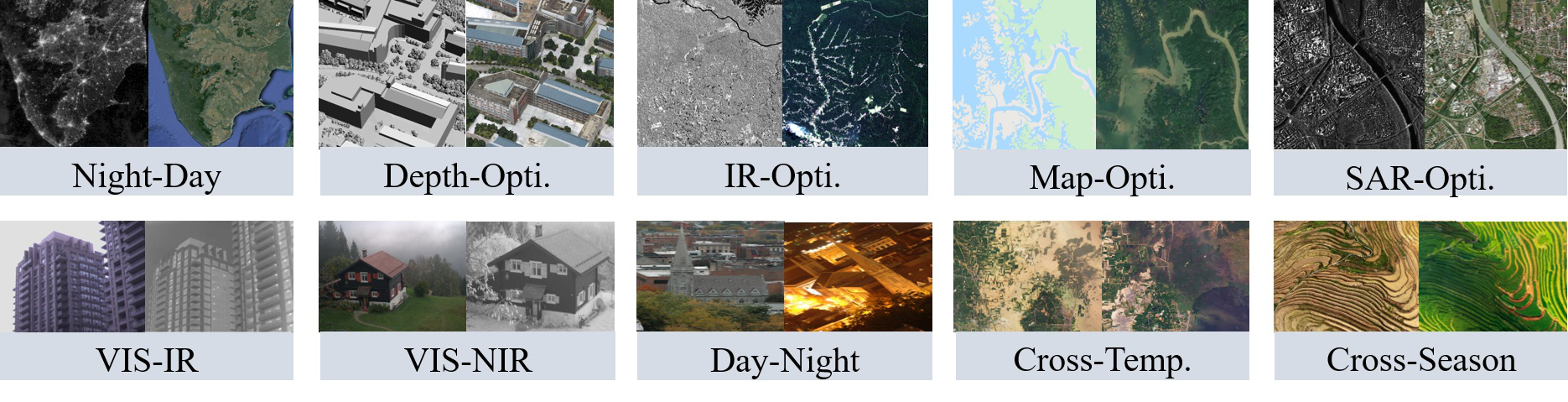}}
 	\caption{Examples of remote sensing images in Multimodal Image Matching Dataset}
 	\label{Fig:Multimodal}
 \end{figure}

 
 The Multimodal Image Matching Datasets \citep{jiang_review_2021} consists of a selection of original image pairs from datasets related to the fields of medicine, remote sensing, and computer vision, involving 164 pairs of multimodal images of 18 different types. \autoref{Fig:Multimodal} presents examples related to remote sensing images in this dataset.
 
\begin{figure}[htbp]
 	\centering
 	\begin{minipage}[t]{0.48\textwidth}
 		\centering
 		\includegraphics[width=\textwidth]{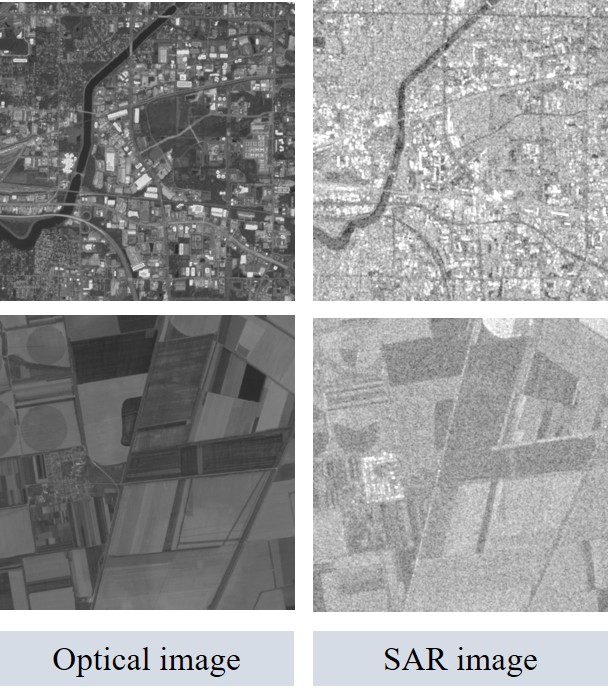}
 		\caption{Examples of SOPatch dataset}
 		\label{Fig:SOPatch}
 	\end{minipage}
 	\hfill
 	\begin{minipage}[t]{0.48\textwidth}
 		\centering
 		\includegraphics[width=\textwidth]{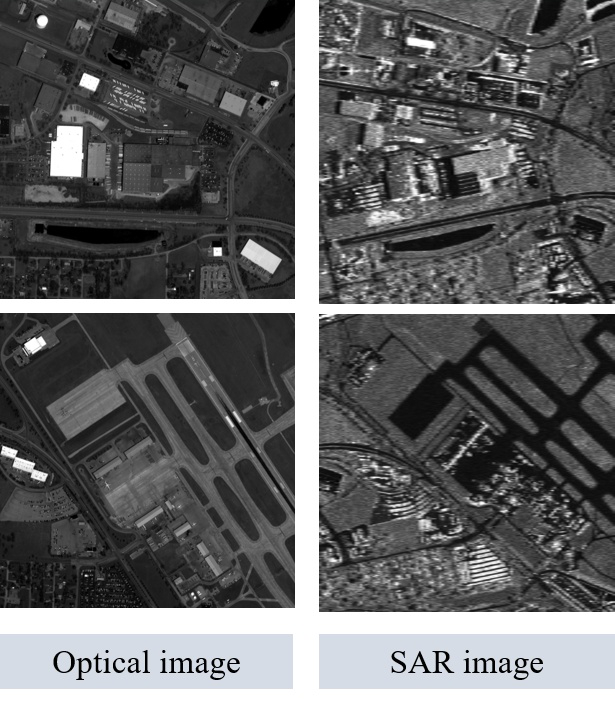}
 		\caption{Examples of OSEval dataset}
 		\label{Fig:OSEval}
 	\end{minipage}
 \end{figure}

 The SOPatch \citep{xu_sar-optical_2023} dataset is built upon the WHU-SEN-City \citep{wang_sar--optical_2019}, OSdataset, and SEN1-2 datasets, undergoing further precise registration to create an optical and SAR image registration dataset containing over 650,000 pairs. This dataset is composed of various satellite images (Sentinel-1, Sentinel-2, Gaofen-3, etc.), including rich scenes (mountains, lakes, buildings, farmlands, bare land, etc.), enabling the training of models with good generalization ability. \autoref{Fig:SOPatch} showcases images of some scenes in this dataset.
 
 
 The OSEval dataset \citep{xiang_global--local_2023} includes an evaluation data set composed of hundreds of pairs of sub-meter resolution optical and SAR images, with various perspectives and covering rich scenes, including airports, residential areas, and industrial areas, as well as dense urban areas. This evaluation data set uses corner points with clear structure in optical and SAR images as ground truth to quantitatively assess optical and SAR registration algorithms, and can ensure unbiased performance comparisons across various registration methodologies. \autoref{Fig:OSEval} showcases images of partial scenes from the OSEval dataset.
 
\begin{figure}[h]
 	\centerline{\includegraphics[width=1\linewidth]{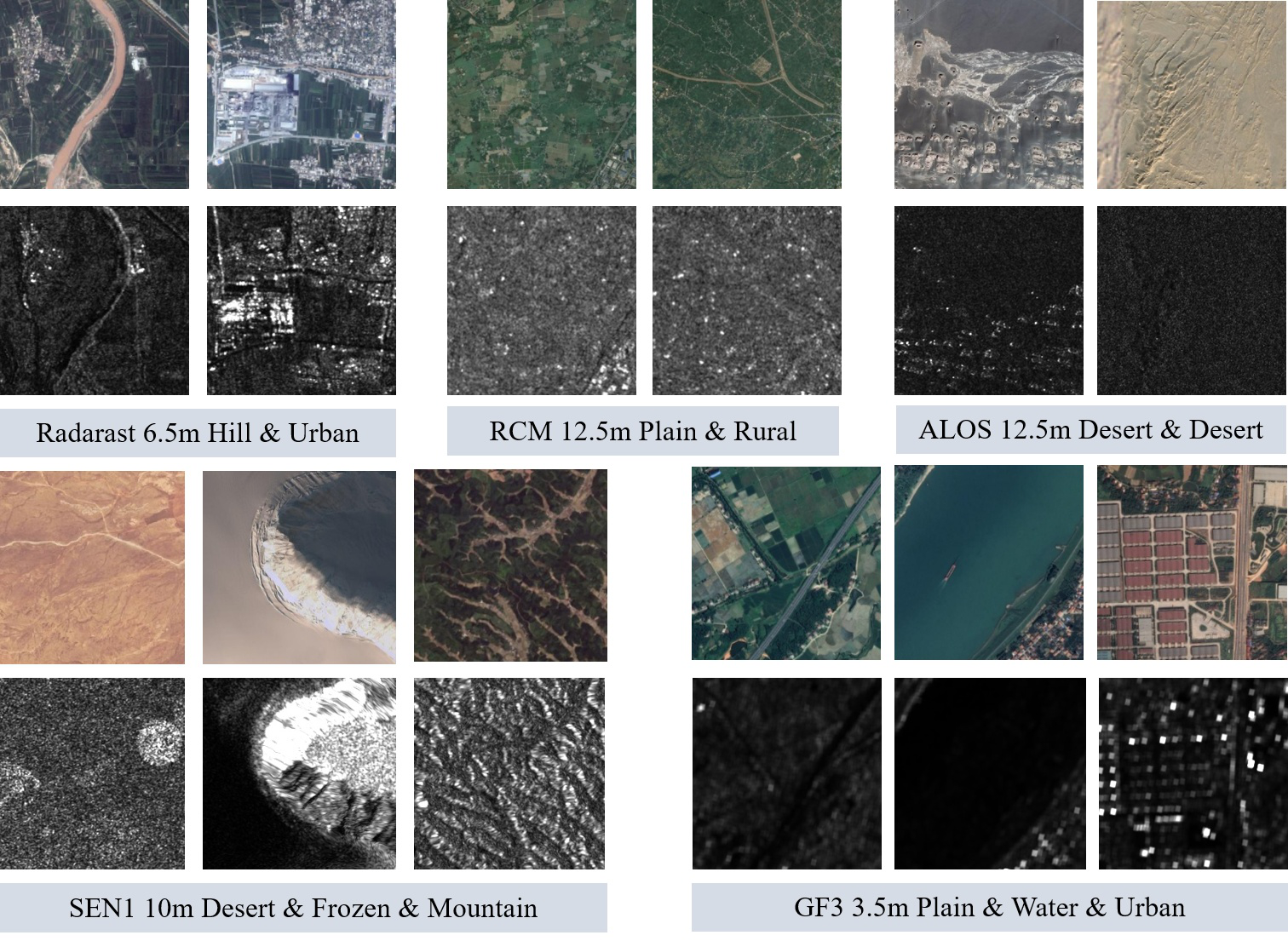}}
 	\caption{Examples of 3MOS dataset}
 	\label{Fig:3mos}
 \end{figure}

 
 3MOS \citep{ye3MOSMultisourceMultiresolution2025} consists of 113,000 pairs of SAR and optical images acquired by five SAR satellites (GF-3, ALOS, Sentinel-1, Radarsat, and RCM) and Google Earth, covering eight distinct scenes including urban, rural, plains, hills, mountains, water, desert, and frozen earth. The spatial resolution of the images ranges from 3.5 meters to 12.5 meters, and the pixel size is 256*256 pixels. \autoref{Fig:3mos} showcases images of partial scenes in the 3MOS dataset.
 
 To systematically evaluate algorithm performance on these datasets, several quantitative metrics are widely adopted. Root Mean Square Error (RMSE) is the primary metric for measuring matching accuracy, as it directly quantifies the residual geometric deviation between point pairs. To assess robustness against severe multimodal discrepancies, the Correct Match Rate (CMR) (or Inlier Ratio) is utilized, reflecting the algorithm's capability to identify true correspondences and reject outliers. Together, these metrics provide a rational and comprehensive evaluation of both precision and reliability.
 
 \section{Summary and prospect of multimodal remote sensing image registration}
 To guide algorithm selection in practical applications, it is essential to weigh the trade-offs between mainstream feature-based methods and modern deep learning approaches. Because feature-based methods are highly interpretable and independent of extensive training data, they are reliable and relatively easy to deploy in scenarios involving significant scale and rotation variations. However, their reliance on handcrafted descriptors leaves them vulnerable to severe nonlinear radiometric discrepancies. Conversely, deep learning methods excel at bridging these complex radiometric gaps and achieving high matching accuracy by learning structural representations directly from the data. Nevertheless, this enhanced performance comes at a cost: deep learning models depend heavily on massive, high-quality datasets, demand significant computational resources (e.g., GPUs) for training and deployment, and often lack theoretical interpretability. Consequently, feature-based methods remain highly practical for applications with limited data or strict hardware constraints, whereas deep learning becomes the preferred choice when robust radiometric alignment is paramount and sufficient computational power is available.
 
 Fundamentally, multimodal image registration serves as a cornerstone for a wide array of downstream remote sensing applications, including image fusion, change detection, and large-scale image stitching. Over the past few decades, research in this domain has shifted from mitigating simple linear radiometric differences to resolving complex nonlinear distortions and severe geometric deformations. By systematically reviewing the three primary methodological paradigms—region-based, feature-based, and deep learning-based methods—this paper provides a comprehensive examination of their theoretical frameworks and representative algorithms. Ultimately, this synthesis aims not only to offer a valuable reference for researchers but also to catalyze further methodological innovation within this rapidly evolving field.
 
 Despite substantial advancements across these paradigms, our review reveals that each category faces inherent constraints, limiting its universal applicability in complex real-world scenarios. Region-based methods, for instance, remain heavily dependent on prior knowledge—such as RPC or POS data, coarse mapping models, and manual tie points—to constrain the initial search space. Furthermore, their performance is acutely sensitive to the overlap area; insufficient coverage frequently leads to registration failure. Alternatively, feature-based methods leverage local spatial relationships to construct high-dimensional descriptors. While effective, this approach often incurs a substantial computational burden and remains susceptible to significant outliers. Such mismatches are particularly prevalent in cross-modal scenarios where scale, rotation, and radiometric variations coexist, sometimes rendering feature-based approaches less robust than their region-based counterparts. Meanwhile, although deep learning-based methods excel at extracting abstract semantic information, their development in remote sensing lags behind the progress seen in the broader computer vision and medical imaging communities. The intricate complexity of remote sensing scenes, combined with high acquisition costs and confidentiality constraints, has led to a scarcity of large-scale, representative datasets. This data bottleneck fundamentally restricts the generalization and evolutionary potential of current learning-based frameworks.
 
 Looking forward, the rapid advancement of integrated sky-ground stereoscopic observation technologies has intensified the demand for intelligent, high-precision, and real-time multimodal registration. Specifically, there is an urgent need to deploy robust registration algorithms in complex, emerging application scenarios. These include rapid post-disaster deformation assessment for emergency response, cross-season temporal registration for long-term ecological monitoring, and high-precision alignment in extreme environments. Addressing these highly heterogeneous and dynamic environments will undoubtedly serve as a primary catalyst for future algorithmic breakthroughs.
 
 To meet the rigorous demands of these advanced applications, future research must evolve across all methodological paradigms. For region-based methods, priority should be given to developing deformation-invariant similarity metrics that integrate local appearance with structural features. Simultaneously, exploring lightweight binary descriptors and hardware-software co-design will be crucial for real-time on-board satellite or airborne processing. For feature-based approaches, the focus must shift toward enhancing computational efficiency—such as by utilizing Fast Fourier Transform acceleration strategies—and implementing more robust outlier rejection algorithms to guarantee consistent sub-pixel precision. Finally, overcoming the data scarcity bottleneck in learning-based paradigms requires not only the curation of more comprehensive public datasets via open-source platforms but also a deeper exploration of unsupervised learning and domain adaptation. Ultimately, by seamlessly integrating geoscientific remote sensing knowledge with advanced computer vision architectures, researchers can design task-specific semantic extraction models tailored to the unique physical mechanisms of multimodal sensors, thereby paving the way for highly resilient and universally applicable registration solutions.


\bibliographystyle{PERS}
\bibliography{references}

\end{document}